%% file: accv2020cameraready.tex
\documentclass[runningheads]{llncs}
\usepackage{graphicx}
\usepackage{amsmath,amssymb} 
\usepackage{color}
\usepackage{tikz}
\usepackage{pgfplots}
\usepackage{comment}
\usepackage{url}
\usepackage{caption}
\usepackage{subcaption}

\newcommand{\etal}{\textit{et. al.}}

\begin{document}
\pagestyle{headings}
\mainmatter

\def\ACCV20SubNumber{556}  

\title{Active Learning for Video Description With Cluster-Regularized Ensemble Ranking} 
\titlerunning{Cluster-Regularized Active Learning}
%
\author{David M. Chan\inst{1}\and
Sudheendra Vijayanarasimhan\inst{2}\and
David A. Ross\inst{2}\and
John F. Canny\inst{1,2}}
\authorrunning{Chan et al.}
%
\institute{University of California, Berkeley, USA\\
\email{\{daviCamera Ready (Copy)dchan,canny\}@berkeley.edu}\and Google Research, USA\\
\email{\{svnaras,dross\}@google.com}}

\maketitle

\begin{abstract}
Automatic video captioning aims to train models to generate text descriptions for all segments in a video, however, the most effective approaches require large amounts of manual annotation which is slow and expensive. Active learning is a promising way to efficiently build a training set for video captioning tasks while reducing the need to manually label uninformative examples. In this work we both explore various active learning approaches for automatic video captioning and show that a cluster-regularized ensemble strategy provides the best active learning approach to efficiently gather training sets for video captioning. We evaluate our approaches on the MSR-VTT and LSMDC datasets using both transformer and LSTM based captioning models and show that our novel strategy can achieve high performance while using up to 60\% fewer training data than the strong state of the art baselines.
\end{abstract}

\section{Introduction}

Automatic video captioning is an emerging area in computer vision research that aims to generate textual descriptions of the visual components of a video. This has various applications including improving video accessibility for the blind and visually impaired~\cite{youdescribe}, summarizing video~\cite{zhang16}, searching and indexing. Unfortunately, training models to do video captioning requires manual descriptions of every second of the video from a large corpus of representative videos. One of the largest current single-clip video captioning datasets, MSR-VTT, has only tens of thousands of unqiue uncorrelated videos whereas solving video captioning will likely require several orders of magnitude more to express the wide diversity of subjects, situations, and relationships possible in video data.

Active learning is a valuable approach in domains where unlabeled and partially labeled examples are readily available but obtaining manual annotations is expensive, such as is the case with automatic video captioning. However, while there has been significant investigation of active learning for computer vision tasks such as object recognition~\cite{collins08}, object detection~\cite{vijayanarasimhan14}, video classification~\cite{yan03} and video segmentation~\cite{vij12}, video captioning has received comparatively little attention. The reason for this is likely rooted in the complexity of the label space. Video captioning requires both sequential input and output, dramatically increasing the complexity of traditional active learning frameworks. To our knowledge, this is one of the first works to define active learning strategies for efficiently collecting training sets for automatic video captioning.

In this paper we explore several active learning strategies for sequence to sequence active learning in video captioning, including uncertainty sampling based on label confidence, sequence entropy and query by committee methods. There are several unique challenges to active learning for deep sequence to sequence models: While traditional active learning methods \cite{settles2009active} select one example at a time to label, retraining the model in its entirety after each new example selection, this strategy is impractical for training models such as transformer networks and LSTMs~\cite{zhou18,venugopalan15iccv}, due to increased training time (hours vs. minutes) and increased inference time (seconds vs. milliseconds). Thus, it is far more efficient to select a large batch of examples at a time to label when using a crowd-sourced collection process \cite{xu2016msr,deng2009imagenet}. Traditional batch-active learning often uses ranking functions which are intractable in deep sequence to sequence learning \cite{hoi09,brinker03,sudheendra:cvpr2010}, making active learning for video description a challenging problem, with no tractable solutions for deep neural networks.

In this work we conduct a thorough empirical analysis of various active learning strategies on two recent and standard video captioning datasets, MSR-VTT and LSMDC, using both transformer based and LSTM based captioning models, and describe a novel cluster-regularized method which is both tractable to compute, and provides strong performance in our test scenario. Our key contributions are:
\begin{enumerate}
    \item Demonstrating that traditional uncertainty sampling techniques do not significantly outperform random sampling, likely because of the difficulty of estimating the sequence entropy.
    \item A novel ensemble based ranking method (Cluster-Regularized Ensemble Divergence Active Learning, Section \ref{sec:methods}) specifically designed for video description active learning which outperform random sampling by a significant margin.
    \item A clustering-based active learning regularization method which can help to increase sample diversity, and when combined with our query-by-committee methods can save as much as 60\% of the manual annotation effort while maintaining high performance (Section \ref{sec:clusters}).
\end{enumerate}

\section{Related Work}

In order to reduce human effort when constructing training sets, various active learning strategies have been proposed for computer vision tasks such as object recognition \cite{collins08,sudheendra:cvpr2009}, detection \cite{vijayanarasimhan14}, video classification \cite{yan03} and video segmentation \cite{vij12}. These methods typically select the next example to query for a label based on uncertainty sampling, entropy, or predicting reductions in risk to the underlying model (see \cite{settles2009active} for a comprehensive review). However, active learning for sequence labeling tasks such as automatic video captioning has received litle attention.

In the natural language processing literature, active learning methods have been proposed for actively selecting examples based on uncertainty sampling \cite{culotta05,scheffer01} or query by committee approaches \cite{dagan95}. In \cite{settles08}, the authors provide a thorough analysis of various active learning methods for sequence labeling tasks using conditional random field (CRF) models. Current state-of-the-art video captioning models, however, typically utilize neural network based architectures such as transformer networks \cite{zhou18} or LSTMs \cite{venugopalan15iccv} and very little research exists on how to successfully apply active learning for complex models --- Transformer networks and LSTMs are expensive to train, taking hours to days to converge, compared to shallow linear models or CRFs employed in previous active learning studies (taking only minutes). Therefore querying a single example at a time is inefficient. It is far more efficient to select a large batch of examples at a time to label when using a crowd-sourced collection process as is typically the case \cite{vijayanarasimhan14}.

Batch-mode active learning methods have been proposed for vision and other tasks in \cite{hoi09,brinker03,sudheendra:cvpr2010}. Batch selection requires more than selecting the $N$-best queries at a given iteration because such a strategy does not account for possible overlap in information. Therefore, the selection functions typically try to balance informativeness with a diversity term to avoid querying overlapping examples \cite{brinker03}. In this work, we take cues from \cite{brinker03}, and develop a batch active-learning method for sequence learning, that is regularized using a measure of information diversity (an idea from \cite{brinker03}), but is tuned to be computed efficiently over sequence learning tasks, such as those in \cite{settles2009active}.

In addition to moving to batch sampling, automated video description is unique in that it has multiple possible correct sequence labels. Recent methods are usually based on expected gradient updates \cite{huang2016active} or the entropy of a sample distribution \cite{settles08}, and are unable to account for scenarios where there are multiple correct labels, or there is dynamic underlying label entropy. In addition, these methods often require computing an estimate of expected model updates over the space of possible labels. This estimate can be extremely expensive for sequence learning (which has exponential label space growth), and there's no clear way of sampling from caption spaces without learning a complex language model. 

Among recent methods, Coreset active learning \cite{sener2017active}, uses an integer linear program (or a greedy optimization) to find a lambda-cover over the feature set. By operating at a feature level, Coreset takes advantage of the semantic compression of the model to find sets of unlabeled samples that are useful to the model's prediction. We discuss our method compared to Coreset in Section \ref{sec:coreset}. 

Some recent methods including VAAL \cite{sinha2019variational} and ALISE \cite{deng2018adversarial} have approached active learning from an adversarial perspective. These methods train a discriminator which attempts to determine which samples are labeled and unlabeled, then select the likely unlabeled samples for training. However, they typically require large number of samples to reliably train the discriminator which is unavailable in the beginning of the active learning process. Nonetheless, it would be an interesting future direction to explore adversarial models for active learning on complex latent spaces. Deep Bayesian active learning \cite{gal2017deep} shows some promise, however strong Bayesian networks for multi-modal vision and language problems are still out out of reach for large scale complex datasets.

\section{Methods}

In this work we introduce a new method for sequence active learning for video description, and compare against several baseline algorithms (Those listed below, along with Coreset \cite{sener2017active} active learning and ALISE \cite{deng2018adversarial}). Throughout this section, we refer to a video $v_i$, and its associated set of descriptions $\mathcal{D} = \{c_{1}(v_i)\dots c_{n}(v_i)\}$. A set of descriptions generated by a model $m_j$ is referred to by $\{c_{m_j,1}(v_i) \dots c_{m_j,n}(v_i)\}$. Videos may have multiple descriptions either through multiple-sampling of the model generative distribution, or through multiple ground-truth labels of the same video. The probability distribution $P_{m_j}(c_i)$ is the likelihood of a description $c_i$ under the model $m_j$, and the distribution $\mathcal{P}^{cond}(m_j, c^{k}_i) = P_{m_j}(c_{i}^{k}(v_i)|c_{i}^{k-1}(v_i), \dots ,c_{i}^{0}(v_i))$ is the conditional distribution of the next word $k$ under the model given the previous words in the description.

\subsection{Active Learning Methods}
\label{sec:methods}

\textbf{Random Selection Baseline:} Our baseline method is to sample new data points from the training set uniformly at random. Random selection is a strong baseline. It directly models the data distribution through sampling, placing emphasis on representative data, but not ``novel" data. Trying to sample outside the random distribution is more likely to cause over-sampling of parts of the data (demonstrated in Figures \ref{fig:cluster_performance}), leading to poorer overall validation performance. \\

\noindent\textbf{Maximum Entropy Active Learning:} Traditional methods for active learning \cite{settles2009active} are often entropy based. As a second strong baseline, we present a maximum entropy active learning method in which we rank samples based on a sample of the entropy of the dataset. Unfortunately, given the exponential number of computations that have to be made in the sequence length, the entropy of the entire output distribution is intractable to compute directly. Thus, to approximate the entropy of the description distribution we compute the mean entropy of the word output distributions at each new word along the generation process of a new description of a sample using our current model. Thus, using a candidate model $m$, we sample $K$ candidate sentences for each video, and we select samples which maximize the ranking function:
\begin{equation}
    R(v_i) = \frac{1}{K}\sum_{k=1}^K\sum_{w=1}^{|c_{m,k}(v_i)|} -P_m(c_{m,k}^w(i))\log{P_m(c_{m,k}^w(v_i))}
\end{equation} where $R(v_i)$ is our approximate estimate of the entropy of any given sample's distribution. \\

\noindent\textbf{Minimum Likelihood Active Learning:} In the minimum likelihood active learning scenario, we select samples where the descriptions that the model generates have the lowest log likelihood under the model distribution. Thus, using a candidate model $m$, we sample $K$ candidate sentences for each video, and then choose samples which minimize the ranking function: 
\begin{equation}
   R(v_i) = \frac{1}{K}\sum_{k=1}^K\sum_{w=1}^{|c_{m,k}(v_i)|} \log{P_m(c_{m,k}^w(v_i)|c_{m,k}^{w-1}(v_i)\dots c_{m,k}^{0}(v_i))} 
\end{equation}
Empirically, we find that the minimum likelihood active learning method is a stronger method than the entropy for use in video captioning (See Figure \ref{fig:results_all}), however this measure of uncertainty suffers from the fact that the model may be very confident about its wrong answers, and will be unable to learn effectively when this is the case. Because these very confident wrong answers are never sampled (or are sampled later in the training process), the model is unable to correct for the initially introduced bias. \\

\noindent \textbf{Query By Committee Ensemble Agreement Active Learning:} To help alleviate the issues with single model uncertainty, we introduce the notion of an ensemble agreement active learning ranking based on query by committee methods for traditional active learning \cite{dagan95}. With this method, we sample a set of likely captions from each member of an ensemble of models (using beam search), and compute the mean pairwise likelihood. For an ensemble of $L$ models $\{m_1,\dots,m_L\}$, from each model $m_l$ we sample captions $\{c_{m_i,1}...c_{m_i,K}\}$ for each available unlabeled video.  Our ranking criterion is then to minimize: \begin{equation}R(v_i) = \frac{1}{L(L-1)}\sum_{p=1}^L\sum_{\substack{q=1\\q\neq p}}^L \sum_{k=1}^K\sum_{w=1}^{|c_{m_p,k}(v_i)|} \frac{\log{ \mathcal{P}^{cond}(m_q, c^{w}_{m_p,k}(i))}}{K|c_{m_p,k}(v_i)|}\end{equation} The idea here is to select samples for labeling which have low agreement, as these are the samples have higher uncertainty under our model/training process. In this scenario, we alleviate many of the concerns with models having  high confidence in wrong answers, as this phenomenon tends to be local to particular models, and these highly incorrect answers will have low likelihood under the learned distributions of the other members of the ensemble.  \\

\noindent \textbf{Query By Committee Ensemble Divergence Active Learning (Proposed Method):} While entropy/perplexity measures for active learning have been well explored in the literature \cite{settles2009active}, it is unclear if these measures are correct for the captioning task. Even if the caption distribution for a video has high entropy, meaning there are many possible likely captions (or even many possible correct captions), this high entropy does not mean that the model is unsure of the outcome. Samples that have many possible captions will thus be over-sampled, since any of the generated captions will have fundamentally lower likelihood than a sample with fewer possible captions. In order to avoid this, we present a method, which computes the KL-divergence between the conditional distributions of the ensemble members. Thus, if the models would choose similar words, given similar inputs - we consider the models to be in agreement. Similarly to the above, for an ensemble of $L$ models $\{m_1...m_L\}$, from each model $m_l$ we sample captions $\{c_{m_i,1}...c_{m_i,K}\}$ for each available unlabeled video. We then choose samples which maximize: \begin{equation}
    R(v_i) = \frac{1}{L(L-1)zhou}\sum_{p=1}^L\sum_{\substack{q=1\\q\neq p}}^L \sum_{k=1}^K\frac{D_{KL}\left(P_{m_p}(c_{m_p,k}(v_i))||P_{m_q}(c_{m_p,k}(v_i))\right)}{K}
\end{equation}
Unfortunately, computing the full joint distribution is prohibitively expensive. Thus, instead we restrict the computation of the divergence to the sum of per-word divergences: 
\begin{equation}
    R(v_i) = \frac{1}{L(L-1)}\sum_{p=1}^L\sum_{\substack{q=1\\q\neq p}}^L\sum_{k=1}^K \frac{D(m_p, m_q, c_{m_p,k}(v_i))}{K}
\end{equation} where
\begin{equation}
\begin{split}
    D(m_p, m_q, c_{m_p,k}(v_i)) & = \\ & \frac{\sum\limits_{w=1}^{|c_{m_p,k}(i)|}D_{KL}\left( \mathcal{P}^{cond}(m_p, c^{w}_{m_p,k}(v_i)) ||  \mathcal{P}^{cond}(m_q, c^{w}_{m_p,k}(v_i)\right)}{|c_{m_p,k}(v_i)|}
\end{split}\end{equation} is the per-word KL-divergence along the generation of the description $c_{m_p,k}(v_i)$ in each of the models. Compared to the likelihood method, this model gives a better estimate of the divergence of the distributions learned by the models of the ensemble. This measure is also independent of the sample length, and distribution perplexity, confounding factors when looking only at the likelihood of the samples. 

\subsection{Improving Diversity With Clustering}\label{sec:clusters}

During the training of the initial active learning models, we noticed through a qualitative investigation that models seemed to be over-sampling parts of the training feature space. This was confirmed by running the experiments shown in Figure \ref{fig:cluster_performance}. To help combat this, we enforced a clustering-based diversity criterion. We first performed a k-means clustering of the training data using the mean (across the temporal dimension) of our visual features. We chose $K=N/20$ clusters, where $N$ is the number of training samples in the dataset. See section \ref{sec:results} for a justification for this number of clusters. We then force the active learning algorithm to select at most $\phi$ samples from each cluster, which notably increases diversity. For the experiments in this paper, we found $\phi=3$  to be the best hyper-parameter value, out of $\phi={1,2,3,\dots 10}$. 

\subsection{Comparison with Coreset Active Learning} \label{sec:coreset} While our method shares some significant similarities at a glance to Coreset~\cite{sener2017active} (i.e. we both use delta-covers of a space to regularize the sampling), they have some notable differences. The Coreset method uses the distribution of the feature space, combined with k-centers over the unlabeled data to select a set of samples which should be annotated. This is equivalent to finding a delta cover over the distribution of the data in the unlabeled space. Our proposed method (Ensemble Divergence + Cluster Regularization) uses the uncertainty of the underlying model to compute a score, and then attempts to regularize this score across the data space by enforcing that no two samples are too close together. Our method not only achieves better performance on our sequence learning tasks, but also runs notably quicker than Coreset, which can fail to solve the Integer Linear Program efficiently. It is interesting future work, however outside the scope of this exploratory paper, to explore selecting among Coresets using our uncertainty metric. Figure \ref{fig:results_main} directly compares Coreset and Greedy Coreset with our proposed model on the video description problem.

\subsection{Models}

The goal of this paper is to explore active learning methods across multiple different model structures. In our experiments we use both a transformer-based model based on Zhou \etal ~\cite{zhou2018end}, and the popular S2VT RNN architecture \cite{venugopalan15iccv} (See supplementary materials for details).  Our models are able to achieve performance comparable to state-of-the-art models using vision-only features and without using policy gradients to optimize a downstream metric \cite{venugopalan15iccv,aafaq2019video}. By adding multi-modal features, and direct REINFORCE optimization, you can gain 7-10 CIDEr points over our implementations \cite{aafaq2019video}. However, while there are more complex model pipelines, we chose two very simple architectures to demonstrate the efficacy of active learning, improve iteration time, and decrease the chance of confounding visual effects. We expect the presented methods to transfer to more complex optimization schemes, and powerful architectures given the flexibility of the formulation and our ablation results.

\subsection{Datasets}

We demonstrate the performance of our model on two common video description datasets, MSR-VTT \cite{xu2016msr} and the LSMDC \cite{rohrbach2017movie}. While these methods may apply to video datasets generated using Automated Speech Recognition (HowTo-100M \cite{miech2019howto100m}) or dense captioning tasks (ActivityNet Captions \cite{krishna2017dense}), we focus on pre-clipped videos with high quality descriptive annotations. We refer the reader to the supplementary materials for a description of the datasets in use.

\subsection{Experimental Setup}

\subsubsection{Feature Extraction and Pre-processing:} To avoid conflating the visual representations of the data with the performance of the captioning model, we follow video-captioning convention and pre-extract features from the videos using a Distill-3D (D3D) \cite{stroud2020d3d} model pre-trained on the Kinetics-600 dataset for activity recognition. The videos are resized on the short-edge to 256px, then center-cropped to 256x256. They are down-sampled to 64 frames at 3 frames per second (with cyclic repetition for videos that are too short), and then passed through the D3D model to generate a 7x1024 representational tensor for the video which is used in the captioning process. The text data is tokenized using a sub-word encoding \cite{kudo2018sentencepiece} with a vocabulary size of 8192.

\subsubsection{Training:} Each model is trained in PyTorch \cite{paszke2017automatic} with a batch-size of 512 for 80 epochs. We use the ADAM \cite{kingma2014adam} optimizer with a warm-up learning rate schedule with 1000 steps of warm-up, ranging from $1e^{-6}$ to $1e^{-3}$, then decaying over 250,000 steps to $0$. We run our experiments using 8 Tesla T4 accelerators on Google Cloud Platform, making use of NVIDIA Apex\footnote{\url{https://github.com/NVIDIA/apex}} for mixed-precision fp-16 training.

\subsection{Evaluation:} In all active learning methods, we begin by seeding the method with 5\% of the data, chosen randomly. For a fair comparison, this random slice is shared across each of the active learning methods. We then train an initial classifier for use with our active learning method. When the classifier has converged (achieved minimum loss on a validation dataset, or trained for 80 epochs, whichever comes first), we use the classifier, and the proposed ranking methods (Using a cluster-limit $\phi=3$, and a set of $8$ sampled captions) to select an additional 5\% of the training data. This happens 19 additional times (20 total evaluation points), until all of the training data has been selected. At each step, we run an evaluation of the model to determine the performance. Exploring the active learning process when using larger and smaller batches is interesting future work --- when selecting very few examples, there is more potential benefit, but more computation required, selecting more samples requires less computation, but can be a more difficult task. Exploring ideas in continual learning, where the classifiers are re-initialized with weights from the previous active learning step is also interesting future work, however we found in practice that this does not heavily influence the training process.

During evaluation, we sample 8 candidate sentences with a temperature of 0.8, which are then evaluated using the COCO Captions Evaluation Tools \cite{chen2015microsoft}. For tuning, we use a validation dataset sub-sampled from the training dataset, and we report the results on the official validation dataset (unseen during training/tuning) below. For ensemble-based metrics, we use the mean performance of the ensemble members. For non-ensemble based metrics, we perform multiple runs of active learning, and report the error as a 95\% bootstrapped confidence interval. While the 95\% is somewhat arbitrary, we present the full trajectories, for readers to explore.

\section{Results \& Discussion}
\label{sec:results}

\input{tex_figures/msrvtt_results}
\input{tex_figures/all_results_v1}

Our key results using the transformer architecture on the MSR-VTT dataset are presented in Figure \ref{fig:results_main}. Clearly, we can see that the clustered-divergence outperforms the benchmark models by a wide margin, using about 25\% of the data to achieve a CIDEr score of 0.38 (95\% of max). A full set of results is shown in Figure \ref{fig:results_all} for the methods from Section \ref{sec:methods}. Some additional qualitative results are presented in the supplementary materials.

Our method is highly prone to over-sampling parts of the distribution. To demonstrate over-sampling by examining the performance of our models across multiple clusters. Figure \ref{fig:cluster_performance} shows that enforcing diversity is key to our approach: If we use no clustering, we actually fail to outperform random performance while adding a few clusters allows us to mitigate this effect and adding sufficient clustering allows for significant performance benefits. We can also see the effect of clustering by examining the mean distance to the validation set over the active learning iterations. We can also see from Figure \ref{fig:cluster_performance} that the agreement method alone is unable to efficiently distribute across the validation set, however random and clustered methods achieve similar distribution effects. It's interesting to note, however, that even without the cluster enforcement the agreement metrics select from more visual diversity than the entropy/likelihood methods - leading to better performance (Table \ref{tab:clusters}). The results for a cluster-regularized random selection method are given in Figure \ref{fig:results_all}, however it is not significantly different from random alone, since the random method already samples uniformly from the set of input samples. Figure \ref{fig:ensemble_members} shows that as we increase the number of ensemble members, the performance increased, however there are diminishing returns, as the models begin to capture the same data.

\begin{table}
\centering
\small
\begin{tabular}{c|c}
    \textbf{Method} & \textbf{Mean Number of Clusters Selected/Iteration} \\
    \hline
    Random Selection &  $195.47 \pm 21.2$ \\
    Cluster-Regularized Divergence & $212.5 \pm 14.4$ \\
    Cluster-Regularized Agreement & $202.3 \pm 17.6$ \\
    Cluster-Regularized Entropy & $215.7 \pm 12.8$ \\
    Agreement Only & $181.00 \pm 16.9$ \\
    Entropy & $160.31 \pm 16.4$ \\
    Likelihood & $169.25 \pm 13.7$ \\
\end{tabular}
\caption{\small{Average number of clusters selected per iteration. The random and cluster-normalized methods select from a wider visual variety of samples, while the non-normalized samples select very few clusters on average.}}
\label{tab:clusters}
\end{table}

\input{tex_figures/cluster_combined}
\input{tex_figures/ensemble_members}

We can also see from Figure \ref{fig:results_all} that the ordering of methods can be dependent on the metrics chosen. While our proposed method outperforms all of the baseline methods, it is most helpful under the CIDEr and ROUGE metrics which prefer higher-level descriptions of the scenes. The method helps less for improving metrics that depend on lower-level semantics, such as BLEU and METEOR. We suspect that this is due to the influence of the active learning method on sampling a diverse set of samples - as increasing the sample diversity can help to improve high-level understanding of a scene, while perhaps having detrimental impacts on the language modeling process.

\begin{figure}[ht]
    \centering
    \includegraphics[width=\linewidth]{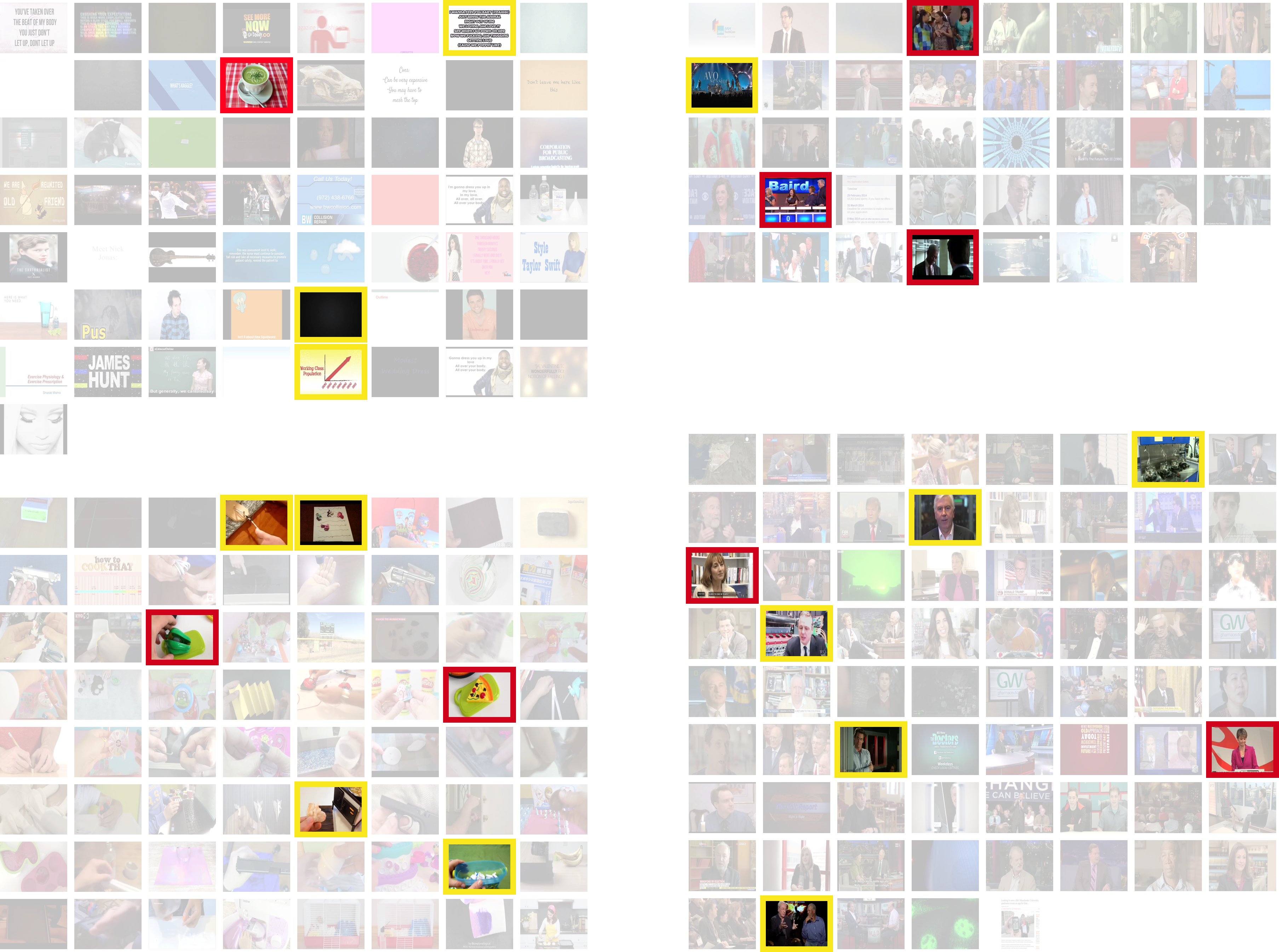}
    \caption{\small{Visualization of four clusters of videos from the training dataset. Highlighted elements were selected by the cluster-divergence learning method (red), or the random method (yellow) in the first two iterations. In clusters with low visual diversity active learning selects fewer samples (top-left, bottom-left, bottom-right), while selecting more samples in clusters with high visual diversity (top-right), suggesting that the active method is choosing more informative samples.}}
    \label{fig:cluster_vis}
\end{figure}

While we have made the case that a strong diversity of samples is required, it is also interesting to look at exactly which samples were selected. Figure \ref{fig:cluster_vis} demonstrates some of the diversity of samples selected by our methods in comparison to the samples selected by the random method. We can see that the active learning method is sampling from a diverse set of elements from each cluster, while the random method is sampling a representative sample, but not necessarily the most relevant or useful videos. 

One important thing to note is that because we are sampling from data that is in the initial training data for the two datasets, the results presented in this paper may be an optimistic upper bound for the performance of an active learning tool. There is a significant amount of cleaning and curating that goes into these datasets which may or may not impact the final results of the classification, and the effort may be higher when annotating video in the wild. Future techniques may need to be developed for efficiently cleaning data, or curating samples that are relevant to captioning as a whole.

One downside to our experimental method is that our models do not achieve optimal performance in each training step, as the optimal hyper-parameters of the model change as more data is added. To ease this issue we use an adaptive training scheme which trains for more iterations than necessary with early stopping, however it is an interesting direction of future work to explore auto-tuning during the learning process to improve performance.

\input{tex_figures/lstm_results}
\input{tex_figures/lsmdc_results}

Our proposed method is not limited to the dataset or model. Figure \ref{fig:results_lsmdc} demonstrates the performance of our best method, clustered divergence, on the LSMDC dataset. We can see here that we achieve a CIDEr score of 0.121 (95\% of max) with only 50\% of data required by random sampling. Thus, we can see that the performance of the active learning method is not just limited to the MSR-VTT dataset. In addition, Figure \ref{fig:results_lstm} demonstrates that the performance is not limited only to our transformer based model. The S2VT model also improves, achieving a CIDEr score of 0.3219 with only 60\% of data required by random selection.

In addition to requiring fewer data, our method can be significantly more efficient than the current state of the art methods. On our test-bench machine, we saw the following ranking times using the MSR-VTT dataset (Samples / Sec): Random: 2012.04, Entropy: 12.41, Cluster-Regularized Ensemble Ranking: 11.08, ALISE: 6.89, Coreset-Optimal: 0.11, and Coreset-Greedy: 11.89.

\section{Conclusion \& Future Work}

In this paper, we have presented an initial set of methods aiming to tackle the active learning problem for video description, a challenging task requiring complex modeling where due to the complexity of the output distribution, many active learning methods are unable to function efficiently, or at all. We have shown that we can achieve 95\% of the full performance of a trained model with between 25\% and 60\% of the training data (and thus, manual labeling effort), across varying models and datasets.  

While pairwise measures among ensemble members may be a good model of uncertainty, there are many such measures. Expected gradient variance methods such as \cite{huang2016active,settles08} are good candidates for future exploration. While such methods now do not account for the complexity of multiple correct labels, and dynamic entropy distributions, we may be able to compute high quality estimates. Such gradient methods may work in scenarios where the KL divergence between the final distributions of the models may be relatively low, but the evaluated sample has useful second-order gradient information.

It is also interesting, and likely fruitful, future work to explore different methods for clustering the elements of the training dataset. In many cases, we would like to enforce a subject-level diversity among the different inputs (as show by Figure \ref{fig:cluster_vis}), however visual similarity may not necessarily be the best metric to use for clustering. Using additional features to rank the diversity of the samples may provide better results, by increasing the individual diversity of each cohort more than k-means clustering in the visual space. 

By exploring the applications of our work in practice, we can build robust active learning methods and collect large and effective datasets for video description. We hope these datasets will be used to improve the performance of downstream description tools in this complex and challenging labeling domain.

\section{Acknowledgements}
This work is supported in part by Berkeley Deep Drive, the BAIR commons project, and Google Cloud Platform. The authors would like to thank Trevor Darrell, Forrest Huang, Roshan Rao, and Philippe Laban for their insightful conversations and assistance.

\clearpage
%
%
\bibliographystyle{splncs}
\bibliography{egbib}

\clearpage
\appendix

\section*{Appendix}

\hfill

\section{Datasets}

We demonstrate the performance of our model on two common video description datasets, MSR-VTT \cite{xu2016msr} and the LSMDC \cite{rohrbach2017movie}.

\subsubsection{MSR-VTT:} The MSR Video to Text Dataset (MSR-VTT) \cite{xu2016msr} is a large-scale benchmark for video description generation. The dataset was generated by collecting a set of 257 popular video queries, selecting 118 videos for each query. These videos were then annotated using Mechanical Turk with 20 natural language sentences. This provides 10K web video clips, with 41.2 hours of video, and 200K clip-description pairs. The clips have an average length of approximately 15 seconds.

\subsubsection{LSMDC:} The Large Scale Movie Description Challenge (LSMDC) \cite{rohrbach2017movie} combines two common bench-mark datasets: M-VAD \cite{torabi2015using} and MPII-MD \cite{rohrbach2015dataset}. The dataset consists of video descriptions extracted from professionally generated Descriptive Video Services tracks on popular movies. The dataset contains 118,081 clips from 202 unique films. Each clip has approximately one sentence of description, are 2-5 seconds each, and the names of most characters are replaced with a signifying "SOMEONE" tag. The LSMDC dataset has a very wide text coverage, with almost 23,000 unique vocabulary tokens.

\clearpage

\section{Models}

We use two models in the paper, specified by Figure \ref{fig:models} below. These models are relatively standard in the Video Description literature. Reference numbers refer to references from the main paper.

\begin{figure}
    \centering
    \begin{subfigure}{\textwidth}
    \centering
    \includegraphics[width=0.8\linewidth]{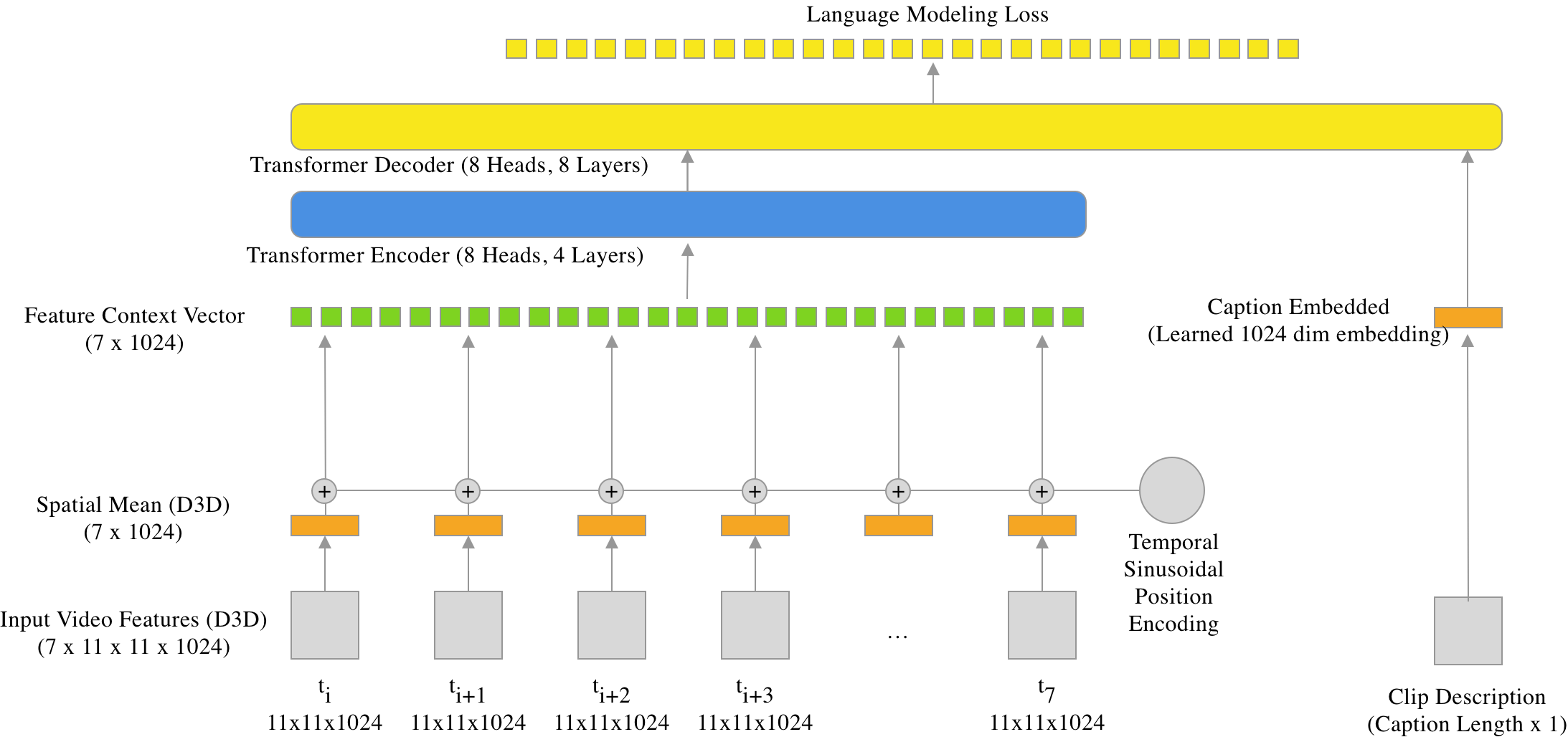}
    \caption{\small{Model architecture of our transformer-based model based on \cite{zhou18}. The only difference in this model and that from \cite{zhou18} is we drop the computation of the video masking which is unnecessary in our task. We use self-attention over spatially pooled input vectors to produce a context for a Transformer Decoder \cite{vaswani2017attention}, which is a conditional cross-attention used to produce the output with a language modeling loss. The best-case performance of this model is similar to the stat-of-the-art method for vision only features presented in \cite{aafaq2019video}.}}
    \end{subfigure}
    \newline
    \begin{subfigure}{\textwidth}
    \centering
    \includegraphics[width=0.8\linewidth]{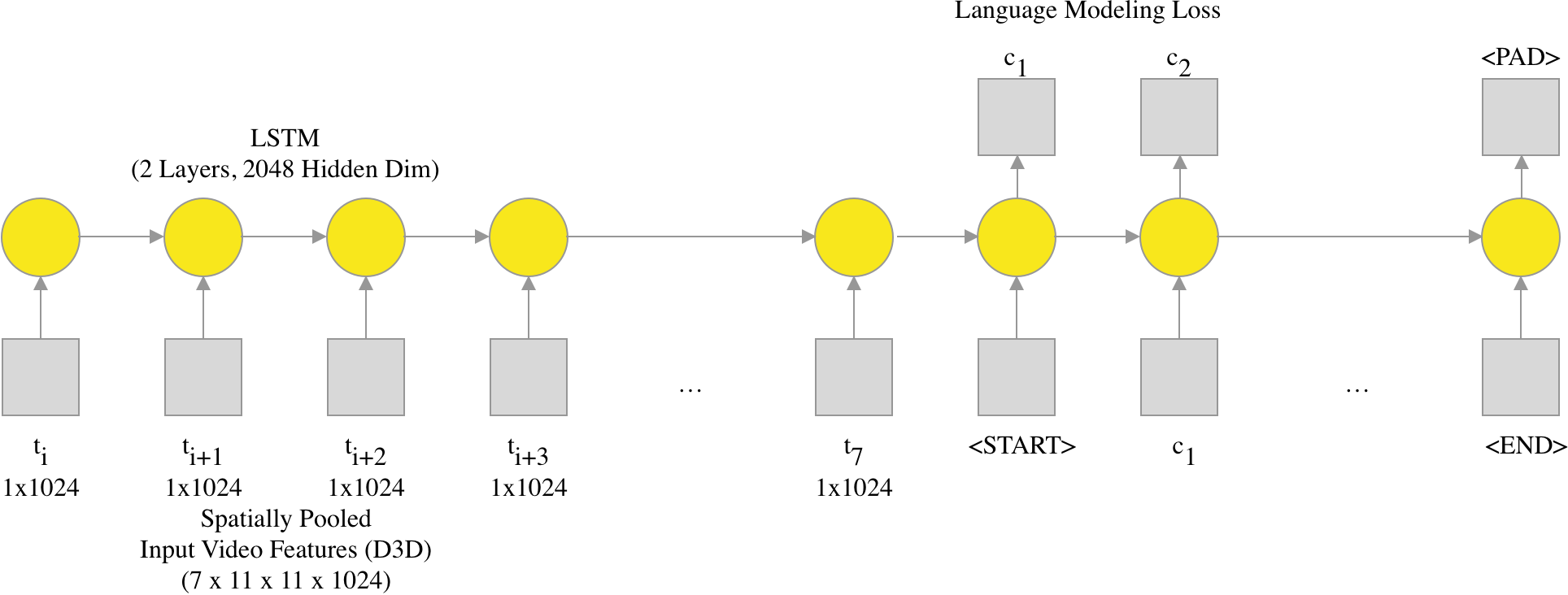}
    \caption{\small{Model architecture of the S2VT model \cite{venugopalan15iccv}. An LSTM encoder is used to encode the spatially pooled video features. The hidden state of the encoder is used to initialize the hidden state of a decoder, which produces the output tokens. The final performance is slightly better than the performance reported in \cite{venugopalan15iccv}.}}
    \end{subfigure}
    \caption{\small{Model diagrams for the two models used in this paper. (a), the transformer-based architecture. (b), the S2VT based architecture.}}
    \label{fig:models}
\end{figure}

\clearpage

\section{Qualitative Examples}

\begin{figure}
    \centering
    \includegraphics[width=0.8\linewidth]{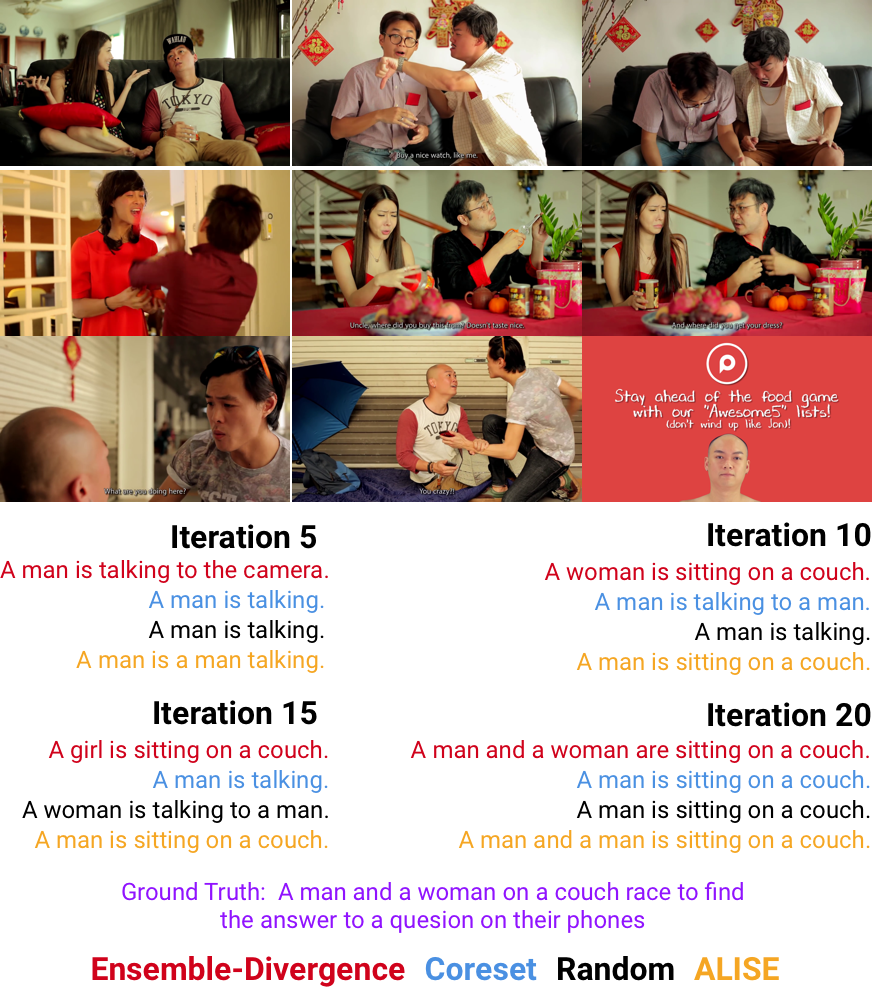}
    \caption{An example description (Selected randomly) traced during the learning process using multiple methods. While this sample does not have a high visual correlation with the ground truth, this is the case for many videos in the MSR-VTT dataset. As we can see, over the course of training the captions evolve, and the proposed method is able to quickly capture the information present in the scene. }
    \label{fig:my_label}
\end{figure}

\clearpage

\end{document}

%% file: tex_figures/msrvtt_results.tex
\begin{figure}[t]
    \centering
    \begin{tikzpicture}
    	\begin{axis}[
    		xlabel=\% of Data Used,
    		ylabel=CIDEr Score,
    		legend cell align=left,
            legend pos=south east,
            legend style={draw=none, nodes={scale=0.7, transform shape}},
            width=\textwidth,
            height=14em,
    		]
    	\addplot[color=red,mark=x, error bars/.cd, y dir=both, y explicit] coordinates {
        	(0.05,0.14459746624285721) +- (0.01820952484705055,0.01820952484705055)
            (0.1,0.34133817502857156) +- (0.011602793800472668,0.011602793800472668)
            (0.2,0.3821888013142857) +- (0.010743159631669286,0.010743159631669286)
            (0.30000000000000004,0.38667965231428564) +- (0.009546186924254484,0.009546186924254484)
            (0.4,0.3875095374571428) +- (0.0005504774246634336,0.0005504774246634336)
            (0.5,0.3907885278857142) +- (0.007487471415951068,0.007487471415951068)
            (0.6000000000000001,0.39385577985714276) +- (0.0049548099338514095,0.0049548099338514095)
            (0.7000000000000001,0.3970387137142857) +- (0.01028215778102504,0.01028215778102504)
            (0.8,0.39939951599999995) +- (0.004900044867840709,0.004900044867840709)
            (0.9,0.4012899259999998) +- (0.010440296640760558,0.010440296640760558)
            (1.0,0.4031686059999996) +- (0.0022299782349606946,0.0022299782349606946)
    	};
    	\addplot[color=green,mark=x, error bars/.cd, y dir=both, y explicit] coordinates {
        	(0.05,0.13765977818498365) +- (0.010511800869859405,0.012264388334717777)
            (0.1,0.23277631562181217) +- (0.001488447917450031,0.01339299551750179)
            (0.2,0.27776768161945786) +- (0.008094248841345988,0.016883796803659673)
            (0.30000000000000004,0.3010329247171738) +- (0.007792850089862591,0.01778047319450717)
            (0.4,0.3319213303264612) +- (0.0003983040181926989,0.008541066236073536)
            (0.5,0.3508618911289101) +- (0.018760080398264375,0.012114019821716107)
            (0.6000000000000001,0.35753657776949815) +- (0.002704373855685267,0.0009245005234609671)
            (0.7000000000000001,0.36229900784933894) +- (0.013887191924642356,0.0010666770837064134)
            (0.8,0.37344831894345853) +- (0.015910488163417252,0.0030340013753234474)
            (0.9,0.38466932687412486) +- (0.016677847431683178,0.007369041376765302)
            (1.0,0.40129757441985836) +- (0.004486251603607503,0.017144318604832)
    	};
    	\addplot[color=blue,mark=x] coordinates {
        	(0.05,0.1482111430907071) +- (0,0)
            (0.1,0.2921352741498232) +- (0,0)
            (0.2,0.3519102051839767) +- (0,0)
            (0.30000000000000004,0.353124756724694) +- (0,0)
            (0.4,0.36841802279066616) +- (0,0)
            (0.5,0.3701687756155284) +- (0,0)
            (0.6000000000000001,0.3733696774311549) +- (0,0)
            (0.7000000000000001,0.381183523964211) +- (0,0)
            (0.8,0.3827769095246234) +- (0,0)
            (0.9,0.3978100768981206) +- (0,0)
            (1.0,0.41052410648807747) +- (0,0)
    	};
    	\addplot[color=orange, mark=x, error bars/.cd, y dir=both, y explicit] coordinates {
        	(0.05, 0.13900874323469123) +- (0.017378783699676863,0.017378783699676863)
            (0.1,0.30512161743025107) +- (0.01712928012891979,0.01712928012891979)
            (0.2,0.34971922760502605) +- (0.011766024602765702,0.011766024602765702)
            (0.30000000000000004,0.35324226525954405) +- (0.01970944913039962,0.01970944913039962)
            (0.4,0.3533087208543085) +- (0.018575543771203688,0.018575543771203688)
            (0.5,0.36263051171303545) +- (0.011277820673133463,0.011277820673133463)
            (0.6000000000000001,0.3829410004413293) +- (6.997099532958462e-05,6.997099532958462e-05)
            (0.7000000000000001,0.3874571484123561) +- (0.009183017033036865,0.009183017033036865)
            (0.8,0.3939557308908091) +- (0.0031679007480770838,0.0031679007480770838)
            (0.9,0.39677706791811757) +- (0.009792803846947981,0.009792803846947981)
            (1.0,0.40004071391928664) +- (0.009972713624096187,0.009972713624096187)
    	};
    	\addplot[color=black, mark=x, error bars/.cd, y dir=both,  y explicit] coordinates {
        	(0.05,0.13537492230488565) +- (0.004142700938563162,0.004142700938563162)
            (0.1,0.30824160358320385) +- (0.004130710522294123,0.004130710522294123)
            (0.2,0.35221427001141226) +- (0.0030704399980735,0.0030704399980735)
            (0.30000000000000004,0.36282498135590796) +- (0.0042113323761944,0.0042113323761944)
            (0.4,0.3725890156538492) +- (0.008621183743958885,0.008621183743958885)
            (0.5,0.37728219702325894) +- (0.011827159577354606,0.011827159577354606)
            (0.6000000000000001,0.3783116852497564) +- (0.018220730573441087,0.018220730573441087)
            (0.7000000000000001,0.3820833092936154) +- (0.01131286143702646,0.01131286143702646)
            (0.8,0.387605776770586) +- (0.00041028147544972085,0.00041028147544972085)
            (0.9,0.3886265215825893) +- (0.016847433503951285,0.016847433503951285)
            (1.0,0.4013012077768124) +- (0.00027061156318226366,0.00027061156318226366)
    	};
    	
    	\addplot[color=black, dashed] coordinates {
    	    (0,.38)
    	    (1,.38)
    	} node[anchor=north east] at (axis description cs: 0.22,0.93) {\scriptsize 95\% of Max};
    	\legend{Ensemble Divergence (Ours), Coreset Greedy \cite{sener2017active}, Coreset \cite{sener2017active}, ALISE \cite{deng2018adversarial}, Random}
    	\end{axis}
    \end{tikzpicture}
    \caption{\small{Validation performance of active learning methods on the MSR-VTT dataset using the CIDEr metric \cite{vedantam2015cider}. Each run represents the mean of a bootstrap sample of ten runs. Our proposed method significantly outperforms all other methods, achieving 95\% of the max performance while using only 25\% of the data. This figure is measured 10 intervals instead of 20, due to the cost of Coreset's ILP solver.}}
    \label{fig:results_main}
\end{figure}

%% file: tex_figures/all_results_v1.tex
\begin{figure}
    \centering
    \begin{subfigure}{\textwidth}
    \begin{tikzpicture}
    	\begin{axis}[
    		ylabel=CIDER Score,
    		width=\textwidth,
            height=13em,
		]
    \addplot[color=green,mark=x,error bars/.cd, y dir=both, y explicit] coordinates {
    		(0.05,0.2326) +- (0.0,0.0)
            (0.1,0.31858) +- (0.0,0.0)
            (0.15,0.33925) +- (0.0,0.0)
            (0.2,0.3503158195165537) +- (0.0,0.0)
            (0.25,0.3619713804040586) +- (0.0,0.0)
            (0.3,0.36974980194252074) +- (0.0,0.0)
            (0.35,0.37022196029791484) +- (0.0,0.0)
            (0.4,0.3694832162562539) +- (0.0,0.0)
            (0.45,0.3696413943267579) +- (0.0,0.0)
            (0.5,0.3698282235223256) +- (0.0,0.0)
            (0.55,0.36402048576683987) +- (0.0,0.0)
            (0.6,0.3634291418218846) +- (0.0,0.0)
            (0.65,0.3690993827602163) +- (0.0,0.0)
            (0.7000000000000001,0.3697310398917909) +- (0.0,0.0)
            (0.75,0.3692674814483858) +- (0.0,0.0)
            (0.8,0.368313531640749) +- (0.0,0.0)
            (0.8500000000000001,0.36855815058563324) +- (0.0,0.0)
            (0.9,0.3720680839191941) +- (0.0,0.0)
            (0.9500000000000001,0.37235969921175987) +- (0.0,0.0)
            (1.0,0.376534954609938) +- (0.0,0.0)
    	};
    	\addplot[color=purple,mark=x,error bars/.cd, y dir=both, y explicit] coordinates {
    		(0.05,0.23410126664193612) +- (0.0,0.0)
            (0.1,0.3048497369356562) +- (0.0,0.0)
            (0.15000000000000002,0.32915335274077195) +- (0.0,0.0)
            (0.2,0.3377061843741477) +- (0.0,0.0)
            (0.25,0.343010072846639) +- (0.0,0.0)
            (0.30000000000000004,0.3499812222499056) +- (0.0,0.0)
            (0.35000000000000003,0.35730064529270267) +- (0.0,0.0)
            (0.4,0.3622567107050786) +- (0.0,0.0)
            (0.45,0.3617348663191654) +- (0.0,0.0)
            (0.5,0.3671089315768593) +- (0.0,0.0)
            (0.55,0.368190088244424) +- (0.0,0.0)
            (0.6000000000000001,0.3677068298011455) +- (0.0,0.0)
            (0.65,0.3694207705960271) +- (0.0,0.0)
            (0.7000000000000001,0.36736032545261776) +- (0.0,0.0)
            (0.75,0.3652674814483858) +- (0.0,0.0)
            (0.8,0.3677399376047741) +- (0.0,0.0)
            (0.8500000000000001,0.36855815058563324) +- (0.0,0.0)
            (0.9,0.3720680839191941) +- (0.0,0.0)
            (0.9500000000000001,0.3682937766050682) +- (0.0,0.0)
            (1.0,0.37640532356934814) +- (0.0,0.0)
    	};
    	\addplot[color=black, dashed, line width=2pt] coordinates {
    		(0.05,0.24115723123809749) +- (0.013089655530755497,0.0138923862569832)
            (0.1,0.2807601868933432) +- (0.006770766136519769,0.006806213276513492)
            (0.15000000000000002,0.30564417784735065) +- (0.009274445881548568,0.009905951294126398)
            (0.2,0.3127827441303963) +- (0.011843520639098609,0.012391686914742761)
            (0.25,0.325228337520881) +- (0.00724654491933452,0.007527101663859289)
            (0.30000000000000004,0.33502442225016893) +- (0.011268830626727747,0.01149540274116131)
            (0.35000000000000003,0.3295700280101429) +- (0.011912141106795459,0.012109035631855678)
            (0.4,0.3369482079796254) +- (0.005259140315692931,0.00557060999222741)
            (0.45,0.3418220671500454) +- (0.013283717278769529,0.013771626866111561)
            (0.5,0.3448165299451313) +- (0.01071709291236975,0.011054167665195905)
            (0.55,0.35025251853541073) +- (0.004907559340565209,0.004980543202449182)
            (0.6000000000000001,0.35341294598274925) +- (0.0024942954911775605,0.002538206795724208)
            (0.65,0.35749911841918114) +- (0.0077876107031587,0.007904299295982561)
            (0.7000000000000001,0.3551895117223253) +- (0.0041259647668138045,0.004192156849029958)
            (0.75,0.35896006144040504) +- (0.0033199583051125724,0.003611053264438136)
            (0.8,0.36392037899856877) +- (0.006705859608862794,0.007342536882400806)
            (0.8500000000000001,0.37298576626320705) +- (0.0028596857217169785,0.002942924032531293)
            (0.9,0.37346207814992054) +- (0.006940521053014359,0.006984639490820473)
            (0.9500000000000001,0.3738013471945252) +- (0.007688572676029563,0.00790992851056391)
            (1.0,0.3756358775437451) +- (0.0049883162491303375,0.0050648910730082645)
    	};
    	\addplot[color=blue,mark=o,error bars/.cd, y dir=both, y explicit] coordinates {
    	    (0.05, 0.2355098708340697) +- (0.0,0.0)
            (0.1, 0.281425736154405) +- (0.0,0.0)
            (0.15, 0.30614877172515687) +- (0.0,0.0)
            (0.2, 0.31362429039155193) +- (0.0,0.0)
            (0.25, 0.33247209185538906) +- (0.0,0.0)
            (0.3, 0.335833063935866) +- (0.0,0.0)
            (0.35, 0.34170707560015355) +- (0.0,0.0)
            (0.4, 0.3448804835216095) +- (0.0,0.0)
            (0.45, 0.348428998785652) +- (0.0,0.0)
            (0.5, 0.34992199664245565) +- (0.0,0.0)
            (0.55, 0.3514256831155741) +- (0.0,0.0)
            (0.6, 0.3528611720896133) +- (0.0,0.0)
            (0.65, 0.3541019546156619) +- (0.0,0.0)
            (0.7, 0.35519470455201657) +- (0.0,0.0)
            (0.75, 0.35822044282843485) +- (0.0,0.0)
            (0.8, 0.36791540100216624) +- (0.0,0.0)
            (0.85, 0.3710640992623989) +- (0.0,0.0)
            (0.9, 0.3741676758842158) +- (0.0,0.0)
            (0.95, 0.37363435336566325) +- (0.0,0.0)
            (1.0, 0.3721470240766391) +- (0.0,0.0)
        };
        \addplot[color=green,mark=o] coordinates {
            (0.05,0.2423889028961552) +- (0,0)
            (0.1,0.3177340007725711) +- (0,0)
            (0.15000000000000002,0.3333782164605046) +- (0,0)
            (0.2,0.3415193558631533) +- (0,0)
            (0.25,0.3529583200393814) +- (0,0)
            (0.30000000000000004,0.3535204392132196) +- (0,0)
            (0.35000000000000003,0.3579443159113561) +- (0,0)
            (0.4,0.35922404603016355) +- (0,0)
            (0.45,0.363642670485712) +- (0,0)
            (0.5,0.3693001754602514) +- (0,0)
            (0.55,0.3694672640131331) +- (0,0)
            (0.6000000000000001,0.3653671492996132) +- (0,0)
            (0.65,0.3663221687092015) +- (0,0)
            (0.7000000000000001,0.367088983710954023) +- (0,0)
            (0.75,0.3676703385956278) +- (0,0)
            (0.8,0.367465733438517653) +- (0,0)
            (0.8500000000000001,0.3684463433678409) +- (0,0)
            (0.9,0.3708061850970226) +- (0,0)
            (0.9500000000000001,0.37156943090920298) +- (0,0)
            (1.0,0.38155265329939975) +- (0,0)
        };
        \addplot[color=purple, mark=o] coordinates {
            (0.05,0.2336041922198925) +- (0,0)
            (0.1,0.2669321810218954) +- (0,0)
            (0.15000000000000002,0.2966060917473622) +- (0,0)
            (0.2,0.3162304834173863) +- (0,0)
            (0.25,0.3216325730759363) +- (0,0)
            (0.30000000000000004,0.32815972434965405) +- (0,0)
            (0.35000000000000003,0.3354257858133343) +- (0,0)
            (0.4,0.3440820436214528) +- (0,0)
            (0.45,0.347417243308272) +- (0,0)
            (0.5,0.34726889516455795) +- (0,0)
            (0.55,0.3500081193882532) +- (0,0)
            (0.6000000000000001,0.35321750627426305) +- (0,0)
            (0.65,0.35985864370367027) +- (0,0)
            (0.7000000000000001,0.3611071127277123) +- (0,0)
            (0.75,0.365912314870499) +- (0,0)
            (0.8,0.368595491899883085) +- (0,0)
            (0.8500000000000001,0.3712513852332387) +- (0,0)
            (0.9,0.3751622038423959) +- (0,0)
            (0.9500000000000001,0.3768779085518753) +- (0,0)
            (1.0,0.377662331637616092) +- (0,0)
        };
        \addplot[color=blue,mark=x] coordinates {
    	    (0.05,0.23358807469535932) +- (0,0)
            (0.1,0.2758549231079982) +- (0,0)
            (0.15000000000000002,0.2880394393204002) +- (0,0)
            (0.2,0.29611980710844727) +- (0,0)
            (0.25,0.30743320979617502) +- (0,0)
            (0.30000000000000004,0.3133236472361788) +- (0,0)
            (0.35000000000000003,0.3176621959417205) +- (0,0)
            (0.4,0.3232477061585097) +- (0,0)
            (0.45,0.3300530081697144) +- (0,0)
            (0.5,0.3378737256531124) +- (0,0)
            (0.55,0.34810311305542069) +- (0,0)
            (0.6000000000000001,0.3527195431222207) +- (0,0)
            (0.65,0.35546019829805038) +- (0,0)
            (0.7000000000000001,0.3542614554274949) +- (0,0)
            (0.75,0.3570648240099439) +- (0,0)
            (0.8,0.3648786060388311) +- (0,0)
            (0.8500000000000001,0.365361154806120658) +- (0,0)
            (0.9,0.36542464461716706) +- (0,0)
            (0.9500000000000001,0.3702450352624367) +- (0,0)
            (1.0,0.378670307069035) +- (0,0)
        };
        \addplot[color=red,mark=x] coordinates {
    	    (0.05,0.2303838330745076) +- (0,0)
            (0.1,0.284014777717478175) +- (0,0)
            (0.15000000000000002,0.292028340078657) +- (0,0)
            (0.2,0.3043926947418053) +- (0,0)
            (0.25,0.30797016810093151) +- (0,0)
            (0.30000000000000004,0.3104511669023759) +- (0,0)
            (0.35000000000000003,0.313535378352247) +- (0,0)
            (0.4,0.3257448067122285) +- (0,0)
            (0.45,0.3309358238487496) +- (0,0)
            (0.5,0.34133449395702104) +- (0,0)
            (0.55,0.34813598348344212) +- (0,0)
            (0.6000000000000001,0.35777629879375095) +- (0,0)
            (0.65,0.3605124243371597) +- (0,0)
            (0.7000000000000001,0.3627146286474699) +- (0,0)
            (0.75,0.3665840025874591) +- (0,0)
            (0.8,0.37004915592306883) +- (0,0)
            (0.8500000000000001,0.3791795732546085) +- (0,0)
            (0.9,0.3807937184292069) +- (0,0)
            (0.9500000000000001,0.38190550301406557) +- (0,0)
            (1.0,0.38152369425658499) +- (0,0)
        };
    \end{axis}
    \end{tikzpicture}
    \end{subfigure}
    \begin{subfigure}{\textwidth}
    \begin{tikzpicture}
    	\begin{axis}[
    		ylabel=METEOR Score,
            width=\textwidth,
            height=13em,
    		]
    	\addplot[color=black, dashed, line width=2pt] coordinates {
        	(0.05,0.23422035625239087) +- (0.0,0.0)
            (0.1,0.23664779350663062) +- (0.0,0.0)
            (0.15000000000000002,0.24175112781949315) +- (0.0,0.0)
            (0.2,0.24709053207804188) +- (0.0,0.0)
            (0.25,0.24743262016793766) +- (0.0,0.0)
            (0.30000000000000004,0.24808543792233087) +- (0.0,0.0)
            (0.35000000000000003,0.25204093798192106) +- (0.0,0.0)
            (0.4,0.25266498702592555) +- (0.0,0.0)
            (0.45,0.2527787060977543) +- (0.0,0.0)
            (0.5,0.25304818821706115) +- (0.0,0.0)
            (0.55,0.2549400741782641) +- (0.0,0.0)
            (0.6000000000000001,0.2558679335763256) +- (0.0,0.0)
            (0.65,0.2559439286265276) +- (0.0,0.0)
            (0.7000000000000001,0.25634285601595225) +- (0.0,0.0)
            (0.75,0.2570689408994138) +- (0.0,0.0)
            (0.8,0.2578526168320202) +- (0.0,0.0)
            (0.8500000000000001,0.2585426008604882) +- (0.0,0.0)
            (0.9,0.260501163380365) +- (0.0,0.0)
            (0.9500000000000001,0.2605901541188439) +- (0.0,0.0)
            (1.0,0.26325242361465195) +- (0.0,0.0)
    	};
    	\addplot[color=blue, mark=o] coordinates {
        	(0.05,0.2339019021915479) +- (0,0)
            (0.1,0.23634439426602566) +- (0,0)
            (0.15000000000000002,0.24051911556368545) +- (0,0)
            (0.2,0.2487816772606242) +- (0,0)
            (0.25,0.248135519270504) +- (0,0)
            (0.30000000000000004,0.2489317245680774) +- (0,0)
            (0.35000000000000003,0.2490807480834559) +- (0,0)
            (0.4,0.2577702737705351) +- (0,0)
            (0.45,0.2572103207234569) +- (0,0)
            (0.5,0.2577333677567312) +- (0,0)
            (0.55,0.2574372181616952) +- (0,0)
            (0.6000000000000001,0.258381580189436627) +- (0,0)
            (0.65,0.2594918896824913) +- (0,0)
            (0.7000000000000001,0.258096428243359) +- (0,0)
            (0.75,0.2582037902452943) +- (0,0)
            (0.8,0.2583348866503715) +- (0,0)
            (0.8500000000000001,0.25942645287856453) +- (0,0)
            (0.9,0.2613765291819378) +- (0,0)
            (0.9500000000000001,0.2601494135982047) +- (0,0)
            (1.0,0.2630518514109085) +- (0,0)
    	};
    	\addplot[color=blue, mark=x] coordinates {
            (0.05,0.2338141045367902) +- (0.0,0.0)
            (0.1,0.23934650528159507) +- (0.0,0.0)
            (0.15000000000000002,0.24664471806433247) +- (0.0,0.0)
            (0.2,0.2507967447122672) +- (0.0,0.0)
            (0.25,0.25166572997584014) +- (0.0,0.0)
            (0.30000000000000004,0.2528006308808832) +- (0.0,0.0)
            (0.35000000000000003,0.2530776838024983) +- (0.0,0.0)
            (0.4,0.2538213347462469) +- (0.0,0.0)
            (0.45,0.25413197787720565) +- (0.0,0.0)
            (0.5,0.25448124595697463) +- (0.0,0.0)
            (0.55,0.2545168527431597) +- (0.0,0.0)
            (0.6000000000000001,0.25613109061884) +- (0.0,0.0)
            (0.65,0.256208781051491) +- (0.0,0.0)
            (0.7000000000000001,0.2565013184536345) +- (0.0,0.0)
            (0.75,0.25707152336932815) +- (0.0,0.0)
            (0.8,0.2580518390108601) +- (0.0,0.0)
            (0.8500000000000001,0.25832694231652675) +- (0.0,0.0)
            (0.9,0.259207548122982) +- (0.0,0.0)
            (0.9500000000000001,0.2605620397435934) +- (0.0,0.0)
            (1.0,0.262913854113611) +- (0.0,0.0)
    	};
    	\addplot[color=red, mark=x] coordinates {
            (0.05,0.2345677553106246) +- (0.0,0.0)
            (0.1,0.24474120051949105) +- (0.0,0.0)
            (0.15000000000000002,0.2480642316493606) +- (0.0,0.0)
            (0.2,0.2518322742843756) +- (0.0,0.0)
            (0.25,0.25219007516713926) +- (0.0,0.0)
            (0.30000000000000004,0.2539589728714434) +- (0.0,0.0)
            (0.35000000000000003,0.2550887747271261) +- (0.0,0.0)
            (0.4,0.25589439058287683) +- (0.0,0.0)
            (0.45,0.25742521681060737) +- (0.0,0.0)
            (0.5,0.2576203795634583) +- (0.0,0.0)
            (0.55,0.25787909569663814) +- (0.0,0.0)
            (0.6000000000000001,0.2597816034891196) +- (0.0,0.0)
            (0.65,0.2599637226264768) +- (0.0,0.0)
            (0.7000000000000001,0.26028731419915996) +- (0.0,0.0)
            (0.75,0.2607197733182926) +- (0.0,0.0)
            (0.8,0.2609265342931955) +- (0.0,0.0)
            (0.8500000000000001,0.26098963278411486) +- (0.0,0.0)
            (0.9,0.261096035071003) +- (0.0,0.0)
            (0.9500000000000001,0.26299664639784914) +- (0.0,0.0)
            (1.0,0.26354273827538577) +- (0.0,0.0)
    	};
    	\addplot[color=green, mark=o] coordinates {
            (0.05,0.2343005690430619) +- (0.0,0.0)
            (0.1,0.24534547425839126) +- (0.0,0.0)
            (0.15000000000000002,0.25053618517173637) +- (0.0,0.0)
            (0.2,0.25234743649015295) +- (0.0,0.0)
            (0.25,0.25567834458730015) +- (0.0,0.0)
            (0.30000000000000004,0.2559045555817484) +- (0.0,0.0)
            (0.35000000000000003,0.2566589920251967) +- (0.0,0.0)
            (0.4,0.25688813888839857) +- (0.0,0.0)
            (0.45,0.25743866032136575) +- (0.0,0.0)
            (0.5,0.25780783358845294) +- (0.0,0.0)
            (0.55,0.25799820730711753) +- (0.0,0.0)
            (0.6000000000000001,0.258164467448083) +- (0.0,0.0)
            (0.65,0.2584267086937678) +- (0.0,0.0)
            (0.7000000000000001,0.2591190711648057) +- (0.0,0.0)
            (0.75,0.2593930197513956) +- (0.0,0.0)
            (0.8,0.2595930111310887) +- (0.0,0.0)
            (0.8500000000000001,0.26053334168794173) +- (0.0,0.0)
            (0.9,0.26179720162654363) +- (0.0,0.0)
            (0.9500000000000001,0.26237328546519667) +- (0.0,0.0)
            (1.0,0.2624821261888196) +- (0.0,0.0)
    	};
    	\addplot[color=purple, mark=o] coordinates {
            (0.05,0.23385294475493737) +- (0.0,0.0)
            (0.1,0.2449718057467822) +- (0.0,0.0)
            (0.15000000000000002,0.24982981195101478) +- (0.0,0.0)
            (0.2,0.2515285094741729) +- (0.0,0.0)
            (0.25,0.2533254621697284) +- (0.0,0.0)
            (0.30000000000000004,0.25432481740810575) +- (0.0,0.0)
            (0.35000000000000003,0.2551191161143054) +- (0.0,0.0)
            (0.4,0.2560149490168102) +- (0.0,0.0)
            (0.45,0.2562398064157887) +- (0.0,0.0)
            (0.5,0.25763037675349515) +- (0.0,0.0)
            (0.55,0.2585476010245801) +- (0.0,0.0)
            (0.6000000000000001,0.2593423949127165) +- (0.0,0.0)
            (0.65,0.26013119275676877) +- (0.0,0.0)
            (0.7000000000000001,0.26033369476364326) +- (0.0,0.0)
            (0.75,0.26085761702794463) +- (0.0,0.0)
            (0.8,0.2611412709641142) +- (0.0,0.0)
            (0.8500000000000001,0.260314971335157) +- (0.0,0.0)
            (0.9,0.26320100892093) +- (0.0,0.0)
            (0.9500000000000001,0.2631907149437139) +- (0.0,0.0)
            (1.0,0.2648871096932336) +- (0.0,0.0)
    	};
    	\addplot[color=green, mark=x] coordinates {
        	(0.05,0.23248886245276265) +- (0.0,0.0)
            (0.1,0.24579410099155394) +- (0.0,0.0)
            (0.15000000000000002,0.2503422579437994) +- (0.0,0.0)
            (0.2,0.2530145523328686) +- (0.0,0.0)
            (0.25,0.2546053687878985) +- (0.0,0.0)
            (0.30000000000000004,0.2566546262099903) +- (0.0,0.0)
            (0.35000000000000003,0.25727623415319184) +- (0.0,0.0)
            (0.4,0.2585026807247063) +- (0.0,0.0)
            (0.45,0.258713186721897) +- (0.0,0.0)
            (0.5,0.2587252901677343) +- (0.0,0.0)
            (0.55,0.2587337965204502) +- (0.0,0.0)
            (0.6000000000000001,0.2588615659365178) +- (0.0,0.0)
            (0.65,0.2588971438483671) +- (0.0,0.0)
            (0.7000000000000001,0.26016415495326195) +- (0.0,0.0)
            (0.75,0.2604276554129255) +- (0.0,0.0)
            (0.8,0.26185406002764233) +- (0.0,0.0)
            (0.8500000000000001,0.26193135343024804) +- (0.0,0.0)
            (0.9,0.26222492407512665) +- (0.0,0.0)
            (0.9500000000000001,0.2630312324323873) +- (0.0,0.0)
            (1.0,0.26402299951999364) +- (0.0,0.0)
    	};
    	\addplot[color=purple, mark=x] coordinates {
            (0.05,0.23305303173328095) +- (0.0,0.0)
            (0.1,0.24562905741769958) +- (0.0,0.0)
            (0.15000000000000002,0.2506303583625919) +- (0.0,0.0)
            (0.2,0.250800345754048) +- (0.0,0.0)
            (0.25,0.2514318156237888) +- (0.0,0.0)
            (0.30000000000000004,0.2542509280645426) +- (0.0,0.0)
            (0.35000000000000003,0.2550409629918154) +- (0.0,0.0)
            (0.4,0.25554891679838765) +- (0.0,0.0)
            (0.45,0.2557581004971683) +- (0.0,0.0)
            (0.5,0.2558292328696388) +- (0.0,0.0)
            (0.55,0.25688131470306663) +- (0.0,0.0)
            (0.6000000000000001,0.2577702734529674) +- (0.0,0.0)
            (0.65,0.2586055127034014) +- (0.0,0.0)
            (0.7000000000000001,0.2590791589824789) +- (0.0,0.0)
            (0.75,0.259132201366447) +- (0.0,0.0)
            (0.8,0.25942485304496216) +- (0.0,0.0)
            (0.8500000000000001,0.2598160621441753) +- (0.0,0.0)
            (0.9,0.26116248936372854) +- (0.0,0.0)
            (0.9500000000000001,0.2618696220289091) +- (0.0,0.0)
            (1.0,0.2625588795306521) +- (0.0,0.0)
    	};
    	
    	\end{axis}
    \end{tikzpicture}
    \end{subfigure}
    \newline
    \begin{subfigure}{\textwidth}
        \begin{tikzpicture}
    	\begin{axis}[
    		ylabel=BLEU Score,
    		width=\textwidth,
            height=13em,
    		]
    	\addplot[color=black, dashed, line width=2pt] coordinates {
            (0.05,0.7141407206570246) +- (0.0,0.0)
            (0.1,0.7208016863445981) +- (0.0,0.0)
            (0.15000000000000002,0.7271735351325991) +- (0.0,0.0)
            (0.2,0.7315355866160045) +- (0.0,0.0)
            (0.25,0.7326505267165095) +- (0.0,0.0)
            (0.30000000000000004,0.7336285170884027) +- (0.0,0.0)
            (0.35000000000000003,0.7336507222677806) +- (0.0,0.0)
            (0.4,0.7341429765244232) +- (0.0,0.0)
            (0.45,0.7380363527208179) +- (0.0,0.0)
            (0.5,0.7394029785850841) +- (0.0,0.0)
            (0.55,0.7396602954929296) +- (0.0,0.0)
            (0.6000000000000001,0.7396843083007263) +- (0.0,0.0)
            (0.65,0.7398721209436978) +- (0.0,0.0)
            (0.7000000000000001,0.740153043549777) +- (0.0,0.0)
            (0.75,0.7404296417647096) +- (0.0,0.0)
            (0.8,0.7409289263753989) +- (0.0,0.0)
            (0.8500000000000001,0.7414032976378955) +- (0.0,0.0)
            (0.9,0.746079163413182) +- (0.0,0.0)
            (0.9500000000000001,0.7469906112515418) +- (0.0,0.0)
            (1.0,0.7498557282286651) +- (0.0,0.0)
    	};
    	\addplot[color=blue, mark=o] coordinates {
    	    (0.05,0.71494598289255) +- (0,0)
            (0.1,0.7180468752410915) +- (0,0)
            (0.15000000000000002,0.72013482967938356) +- (0,0)
            (0.2,0.7273251496603118) +- (0,0)
            (0.25,0.730534209382137) +- (0,0)
            (0.30000000000000004,0.73319210564196822) +- (0,0)
            (0.35000000000000003,0.73544813750632688) +- (0,0)
            (0.4,0.7381921011594932) +- (0,0)
            (0.45,0.7422769295494882) +- (0,0)
            (0.5,0.74273028795860062) +- (0,0)
            (0.55,0.74613193770512556) +- (0,0)
            (0.6000000000000001,0.7479103951270631) +- (0,0)
            (0.65,0.7482131588064785) +- (0,0)
            (0.7000000000000001,0.7475893039100182) +- (0,0)
            (0.75,0.7472200032787201) +- (0,0)
            (0.8,0.7488407513593356) +- (0,0)
            (0.8500000000000001,0.74825332224276332) +- (0,0)
            (0.9,0.7488822910846404) +- (0,0)
            (0.9500000000000001,0.7489223927750065) +- (0,0)
            (1.0,0.74921393045832562) +- (0,0)
    	};
    	\addplot[color=blue, mark=x] coordinates {
            (0.05,0.7139988190223866) +- (0.0,0.0)
            (0.1,0.7233897989226311) +- (0.0,0.0)
            (0.15000000000000002,0.7259549254079785) +- (0.0,0.0)
            (0.2,0.7312601336335706) +- (0.0,0.0)
            (0.25,0.731512669804262) +- (0.0,0.0)
            (0.30000000000000004,0.7332410230819727) +- (0.0,0.0)
            (0.35000000000000003,0.7344216471351941) +- (0.0,0.0)
            (0.4,0.7344477503230651) +- (0.0,0.0)
            (0.45,0.735114872268755) +- (0.0,0.0)
            (0.5,0.7395859544778582) +- (0.0,0.0)
            (0.55,0.739979765377005) +- (0.0,0.0)
            (0.6000000000000001,0.7405334040288021) +- (0.0,0.0)
            (0.65,0.742000237633792) +- (0.0,0.0)
            (0.7000000000000001,0.7422360023199828) +- (0.0,0.0)
            (0.75,0.7426286396619983) +- (0.0,0.0)
            (0.8,0.7433006839736331) +- (0.0,0.0)
            (0.8500000000000001,0.7457771690189391) +- (0.0,0.0)
            (0.9,0.7460196334477524) +- (0.0,0.0)
            (0.9500000000000001,0.7466014238461499) +- (0.0,0.0)
            (1.0,0.7535842537798574) +- (0.0,0.0)
    	};
    	\addplot[color=red, mark=x] coordinates {
            (0.05,0.7149085953328911) +- (0.0,0.0)
            (0.1,0.730328282002455) +- (0.0,0.0)
            (0.15000000000000002,0.7305513149767242) +- (0.0,0.0)
            (0.2,0.7312584717580382) +- (0.0,0.0)
            (0.25,0.7335200802381419) +- (0.0,0.0)
            (0.30000000000000004,0.7362914868638553) +- (0.0,0.0)
            (0.35000000000000003,0.7363220011085725) +- (0.0,0.0)
            (0.4,0.7367713936484083) +- (0.0,0.0)
            (0.45,0.7397680717454294) +- (0.0,0.0)
            (0.5,0.7401766852455494) +- (0.0,0.0)
            (0.55,0.7402866862894691) +- (0.0,0.0)
            (0.6000000000000001,0.7425582283919523) +- (0.0,0.0)
            (0.65,0.742604593177826) +- (0.0,0.0)
            (0.7000000000000001,0.7427135365850257) +- (0.0,0.0)
            (0.75,0.7429833527296118) +- (0.0,0.0)
            (0.8,0.7433266193652419) +- (0.0,0.0)
            (0.8500000000000001,0.7453632188410705) +- (0.0,0.0)
            (0.9,0.7459727911525279) +- (0.0,0.0)
            (0.9500000000000001,0.7467250859815828) +- (0.0,0.0)
            (1.0,0.7473204660900604) +- (0.0,0.0)
    	};
    	\addplot[color=green, mark=o] coordinates {
            (0.05,0.714275034690569) +- (0.0,0.0)
            (0.1,0.726073958730469) +- (0.0,0.0)
            (0.15000000000000002,0.7339469968479715) +- (0.0,0.0)
            (0.2,0.734934902011257) +- (0.0,0.0)
            (0.25,0.735541392250034) +- (0.0,0.0)
            (0.30000000000000004,0.7357342163603902) +- (0.0,0.0)
            (0.35000000000000003,0.7358758889884803) +- (0.0,0.0)
            (0.4,0.7361161133892675) +- (0.0,0.0)
            (0.45,0.7367430727045854) +- (0.0,0.0)
            (0.5,0.7406061754013613) +- (0.0,0.0)
            (0.55,0.7411907884290203) +- (0.0,0.0)
            (0.6000000000000001,0.7427056129306977) +- (0.0,0.0)
            (0.65,0.7437710291846749) +- (0.0,0.0)
            (0.7000000000000001,0.7442145737319239) +- (0.0,0.0)
            (0.75,0.7462410528293253) +- (0.0,0.0)
            (0.8,0.7470032795438473) +- (0.0,0.0)
            (0.8500000000000001,0.747289385696623) +- (0.0,0.0)
            (0.9,0.7496523277463156) +- (0.0,0.0)
            (0.9500000000000001,0.7508234096443659) +- (0.0,0.0)
            (1.0,0.7510983262443962) +- (0.0,0.0)
    	};
    	\addplot[color=purple, mark=o] coordinates {
            (0.05,0.7145310969604388) +- (0.0,0.0)
            (0.1,0.7190350437884636) +- (0.0,0.0)
            (0.15000000000000002,0.7245248102620331) +- (0.0,0.0)
            (0.2,0.725024149855845) +- (0.0,0.0)
            (0.25,0.730326575827196) +- (0.0,0.0)
            (0.30000000000000004,0.7317009216588275) +- (0.0,0.0)
            (0.35000000000000003,0.7333385897389307) +- (0.0,0.0)
            (0.4,0.7343026638719942) +- (0.0,0.0)
            (0.45,0.739793146971117) +- (0.0,0.0)
            (0.5,0.7406431772957613) +- (0.0,0.0)
            (0.55,0.7423964472899888) +- (0.0,0.0)
            (0.6000000000000001,0.7433284833140569) +- (0.0,0.0)
            (0.65,0.7435537763538694) +- (0.0,0.0)
            (0.7000000000000001,0.7450319222785458) +- (0.0,0.0)
            (0.75,0.74989879306903) +- (0.0,0.0)
            (0.8,0.7491412709641142) +- (0.0,0.0)
            (0.8500000000000001,0.750314971335157) +- (0.0,0.0)
            (0.9,0.75020100892093) +- (0.0,0.0)
            (0.9500000000000001,0.75
            11907149437139) +- (0.0,0.0)
            (1.0,0.7518871096932336) +- (0.0,0.0)
    	};
    	\addplot[color=green, mark=x] coordinates {
            (0.05,0.7142886999079228) +- (0.0,0.0)
            (0.1,0.7260415660094189) +- (0.0,0.0)
            (0.15000000000000002,0.7347714836650778) +- (0.0,0.0)
            (0.2,0.7361865073652205) +- (0.0,0.0)
            (0.25,0.7396271602924984) +- (0.0,0.0)
            (0.30000000000000004,0.7401231412921925) +- (0.0,0.0)
            (0.35000000000000003,0.7402686659390227) +- (0.0,0.0)
            (0.4,0.7408391304040907) +- (0.0,0.0)
            (0.45,0.7413995063918304) +- (0.0,0.0)
            (0.5,0.7455119927617939) +- (0.0,0.0)
            (0.55,0.7457045979506228) +- (0.0,0.0)
            (0.6000000000000001,0.7469779284730251) +- (0.0,0.0)
            (0.65,0.7473459142882082) +- (0.0,0.0)
            (0.7000000000000001,0.7475064620338199) +- (0.0,0.0)
            (0.75,0.748797770097313) +- (0.0,0.0)
            (0.8,0.7495395542135416) +- (0.0,0.0)
            (0.8500000000000001,0.7501214184762489) +- (0.0,0.0)
            (0.9,0.7502704749570207) +- (0.0,0.0)
            (0.9500000000000001,0.750619996915536) +- (0.0,0.0)
            (1.0,0.7507072634723233) +- (0.0,0.0)
    	};
    	\addplot[color=purple, mark=x] coordinates {
            (0.05,0.7143033698477079) +- (0.0,0.0)
            (0.1,0.7264744577857899) +- (0.0,0.0)
            (0.15000000000000002,0.7361600950807802) +- (0.0,0.0)
            (0.2,0.737021519229477) +- (0.0,0.0)
            (0.25,0.7379977555248348) +- (0.0,0.0)
            (0.30000000000000004,0.7383676664028112) +- (0.0,0.0)
            (0.35000000000000003,0.7405663006342755) +- (0.0,0.0)
            (0.4,0.7407751811073131) +- (0.0,0.0)
            (0.45,0.7410164055187987) +- (0.0,0.0)
            (0.5,0.7423642716401824) +- (0.0,0.0)
            (0.55,0.7431541870006166) +- (0.0,0.0)
            (0.6000000000000001,0.7444517820992174) +- (0.0,0.0)
            (0.65,0.7444723388818795) +- (0.0,0.0)
            (0.7000000000000001,0.7460670376565453) +- (0.0,0.0)
            (0.75,0.7469866098025078) +- (0.0,0.0)
            (0.8,0.7489513421799361) +- (0.0,0.0)
            (0.8500000000000001,0.7496254804924594) +- (0.0,0.0)
            (0.9,0.7509568682330494) +- (0.0,0.0)
            (0.9500000000000001,0.7513125655229335) +- (0.0,0.0)
            (1.0,0.751623049755178) +- (0.0,0.0)
    	};
	\end{axis}
    \end{tikzpicture}
    \end{subfigure}
    \vspace{1em}
    \newline
    \begin{subfigure}{\textwidth}
        \begin{tikzpicture}
    	\begin{axis}[
    		ylabel=ROUGE Score,
    		xlabel=\% Of Data Used,
    		width=\textwidth,
            height=16em,
            legend cell align=left,
            legend pos=south east,
            legend style={draw=none, nodes={scale=0.7, transform shape}},
    		]
    	\addplot[color=black, dashed, line width=2pt] coordinates {
        	(0.05,0.5247825550776611) +- (0.0,0.0)
            (0.1,0.529852211332443) +- (0.0,0.0)
            (0.15000000000000002,0.5366709346654868) +- (0.0,0.0)
            (0.2,0.5409957336509104) +- (0.0,0.0)
            (0.25,0.5441001102401122) +- (0.0,0.0)
            (0.30000000000000004,0.5455179409719608) +- (0.0,0.0)
            (0.35000000000000003,0.5478126601702844) +- (0.0,0.0)
            (0.4,0.5499958664900081) +- (0.0,0.0)
            (0.45,0.5521345865281444) +- (0.0,0.0)
            (0.5,0.5526519474933168) +- (0.0,0.0)
            (0.55,0.552942652646533) +- (0.0,0.0)
            (0.6000000000000001,0.5529915714623052) +- (0.0,0.0)
            (0.65,0.5553861498418347) +- (0.0,0.0)
            (0.7000000000000001,0.5554757502877685) +- (0.0,0.0)
            (0.75,0.5560776239957659) +- (0.0,0.0)
            (0.8,0.5561137040686712) +- (0.0,0.0)
            (0.8500000000000001,0.5576766540565435) +- (0.0,0.0)
            (0.9,0.5615572622698368) +- (0.0,0.0)
            (0.9500000000000001,0.5616280164377427) +- (0.0,0.0)
            (1.0,0.5630691442639919) +- (0.0,0.0)
    	};
    	\addplot[color=blue, mark=x] coordinates {
        	(0.05,0.5269048072821495) +- (0.0,0.0)
            (0.1,0.53309261659892) +- (0.0,0.0)
            (0.15000000000000002,0.5463857568242352) +- (0.0,0.0)
            (0.2,0.5465034450626252) +- (0.0,0.0)
            (0.25,0.5477169694730938) +- (0.0,0.0)
            (0.30000000000000004,0.5488722771879923) +- (0.0,0.0)
            (0.35000000000000003,0.5497731593274502) +- (0.0,0.0)
            (0.4,0.5499679745324284) +- (0.0,0.0)
            (0.45,0.5524516909141428) +- (0.0,0.0)
            (0.5,0.5525990454511243) +- (0.0,0.0)
            (0.55,0.5527120061310601) +- (0.0,0.0)
            (0.6000000000000001,0.5545000230890846) +- (0.0,0.0)
            (0.65,0.5558467232058527) +- (0.0,0.0)
            (0.7000000000000001,0.5560687236053701) +- (0.0,0.0)
            (0.75,0.5562439011460771) +- (0.0,0.0)
            (0.8,0.559888771556031) +- (0.0,0.0)
            (0.8500000000000001,0.5604996288811106) +- (0.0,0.0)
            (0.9,0.5615445344304311) +- (0.0,0.0)
            (0.9500000000000001,0.5622537486182053) +- (0.0,0.0)
            (1.0,0.5628353871152452) +- (0.0,0.0)
    	};
    	\addplot[color=red, mark=x] coordinates {
    	    (0.05,0.5250839369347442) +- (0.0,0.0)
            (0.1,0.5389242417349751) +- (0.0,0.0)
            (0.15000000000000002,0.5397076105607308) +- (0.0,0.0)
            (0.2,0.5423663875900527) +- (0.0,0.0)
            (0.25,0.5481977172298533) +- (0.0,0.0)
            (0.30000000000000004,0.5510741138649754) +- (0.0,0.0)
            (0.35000000000000003,0.5515056010124706) +- (0.0,0.0)
            (0.4,0.5519524737697148) +- (0.0,0.0)
            (0.45,0.5533187515777876) +- (0.0,0.0)
            (0.5,0.5550754414636955) +- (0.0,0.0)
            (0.55,0.5557726898422185) +- (0.0,0.0)
            (0.6000000000000001,0.5569436756790647) +- (0.0,0.0)
            (0.65,0.5571773740686625) +- (0.0,0.0)
            (0.7000000000000001,0.5579099854615316) +- (0.0,0.0)
            (0.75,0.5583122041576005) +- (0.0,0.0)
            (0.8,0.5588516844568912) +- (0.0,0.0)
            (0.8500000000000001,0.5602265252622061) +- (0.0,0.0)
            (0.9,0.5611159267516204) +- (0.0,0.0)
            (0.9500000000000001,0.5638951778728887) +- (0.0,0.0)
            (1.0,0.564314847236459) +- (0.0,0.0)
    	};
    	\addplot[color=green, mark=o] coordinates {
        	(0.05,0.5226734798518098) +- (0.0,0.0)
            (0.1,0.5400899218687318) +- (0.0,0.0)
            (0.15000000000000002,0.5419784995194725) +- (0.0,0.0)
            (0.2,0.5504389275265484) +- (0.0,0.0)
            (0.25,0.5511184244630094) +- (0.0,0.0)
            (0.30000000000000004,0.5514525393459466) +- (0.0,0.0)
            (0.35000000000000003,0.5522595825687742) +- (0.0,0.0)
            (0.4,0.5530593871393361) +- (0.0,0.0)
            (0.45,0.5538028626812512) +- (0.0,0.0)
            (0.5,0.5540990302698949) +- (0.0,0.0)
            (0.55,0.554626907390348) +- (0.0,0.0)
            (0.6000000000000001,0.5554802045364065) +- (0.0,0.0)
            (0.65,0.5556769603652707) +- (0.0,0.0)
            (0.7000000000000001,0.5563279028857906) +- (0.0,0.0)
            (0.75,0.5573301876939746) +- (0.0,0.0)
            (0.8,0.5574270837549526) +- (0.0,0.0)
            (0.8500000000000001,0.5581857219293245) +- (0.0,0.0)
            (0.9,0.55923135200206) +- (0.0,0.0)
            (0.9500000000000001,0.5595765826159631) +- (0.0,0.0)
            (1.0,0.5636783579368574) +- (0.0,0.0)
    	};
    	\addplot[color=purple, mark=o] coordinates {
        	(0.05,0.525225236126875) +- (0.0,0.0)
            (0.1,0.5368321130261902) +- (0.0,0.0)
            (0.15000000000000002,0.5433581275950683) +- (0.0,0.0)
            (0.2,0.5464090161329904) +- (0.0,0.0)
            (0.25,0.546602149047587) +- (0.0,0.0)
            (0.30000000000000004,0.5470857202816892) +- (0.0,0.0)
            (0.35000000000000003,0.5511289152458124) +- (0.0,0.0)
            (0.4,0.553612317927477) +- (0.0,0.0)
            (0.45,0.5540273639612459) +- (0.0,0.0)
            (0.5,0.5559088392175899) +- (0.0,0.0)
            (0.55,0.5586583089131445) +- (0.0,0.0)
            (0.6000000000000001,0.5588447893460521) +- (0.0,0.0)
            (0.65,0.5601513240904383) +- (0.0,0.0)
            (0.7000000000000001,0.5606134731808425) +- (0.0,0.0)
            (0.75,0.5608989879306903) +- (0.0,0.0)
            (0.8,0.5621412709641142) +- (0.0,0.0)
            (0.8500000000000001,0.563314971335157) +- (0.0,0.0)
            (0.9,0.56320100892093) +- (0.0,0.0)
            (0.9500000000000001,0.5641907149437139) +- (0.0,0.0)
            (1.0,0.5658871096932336) +- (0.0,0.0)
    	};
    	\addplot[color=green, mark=x] coordinates {
        	(0.05,0.5258616240334652) +- (0.0,0.0)
    	    (0.1,0.5443183722681657) +- (0.0,0.0)
            (0.15000000000000002,0.5500963204874765) +- (0.0,0.0)
            (0.2,0.5523611650027507) +- (0.0,0.0)
            (0.25,0.553353593836112) +- (0.0,0.0)
            (0.30000000000000004,0.5541793650889598) +- (0.0,0.0)
            (0.35000000000000003,0.5563902907781554) +- (0.0,0.0)
            (0.4,0.5567950501387045) +- (0.0,0.0)
            (0.45,0.5577935363161459) +- (0.0,0.0)
            (0.5,0.5583851159874607) +- (0.0,0.0)
            (0.55,0.5584260832216481) +- (0.0,0.0)
            (0.6000000000000001,0.5588574612737924) +- (0.0,0.0)
            (0.65,0.5592138608437325) +- (0.0,0.0)
            (0.7000000000000001,0.5599601758641017) +- (0.0,0.0)
            (0.75,0.559988340396532) +- (0.0,0.0)
            (0.8,0.5601879877786375) +- (0.0,0.0)
            (0.8500000000000001,0.560846934500025) +- (0.0,0.0)
            (0.9,0.5614551429470184) +- (0.0,0.0)
            (0.9500000000000001,0.5619963151374446) +- (0.0,0.0)
            (1.0,0.56423682793726) +- (0.0,0.0)
    	};
    	\addplot[color=purple, mark=x] coordinates {
    	    (0.05,0.5233795138855573) +- (0.0,0.0)
            (0.1,0.5400181468505851) +- (0.0,0.0)
            (0.15000000000000002,0.5460653717854946) +- (0.0,0.0)
            (0.2,0.5472149420535378) +- (0.0,0.0)
            (0.25,0.5481830885624248) +- (0.0,0.0)
            (0.30000000000000004,0.5490066107655355) +- (0.0,0.0)
            (0.35000000000000003,0.552378076125166) +- (0.0,0.0)
            (0.4,0.5526091012307448) +- (0.0,0.0)
            (0.45,0.5526115943724196) +- (0.0,0.0)
            (0.5,0.5536914174982439) +- (0.0,0.0)
            (0.55,0.5537027026300674) +- (0.0,0.0)
            (0.6000000000000001,0.5537558341792664) +- (0.0,0.0)
            (0.65,0.5550010894031414) +- (0.0,0.0)
            (0.7000000000000001,0.5569278019415282) +- (0.0,0.0)
            (0.75,0.5589987063398618) +- (0.0,0.0)
            (0.8,0.5601825873607151) +- (0.0,0.0)
            (0.8500000000000001,0.560622858915728) +- (0.0,0.0)
            (0.9,0.5611620561713804) +- (0.0,0.0)
            (0.9500000000000001,0.5643155657216652) +- (0.0,0.0)
            (1.0,0.5649778867714437) +- (0.0,0.0)
    	};
    	\addplot[color=blue, mark=o] coordinates {
    	    (0.05,0.5233795138855573) +- (0.0,0.0)
            (0.1,0.5400181468505851) +- (0.0,0.0)
            (0.15000000000000002,0.5460653717854946) +- (0.0,0.0)
            (0.2,0.5472149420535378) +- (0.0,0.0)
            (0.25,0.5481830885624248) +- (0.0,0.0)
            (0.30000000000000004,0.5490066107655355) +- (0.0,0.0)
            (0.35000000000000003,0.552378076125166) +- (0.0,0.0)
            (0.4,0.5526091012307448) +- (0.0,0.0)
            (0.45,0.5526115943724196) +- (0.0,0.0)
            (0.5,0.5536914174982439) +- (0.0,0.0)
            (0.55,0.5537027026300674) +- (0.0,0.0)
            (0.6000000000000001,0.5537558341792664) +- (0.0,0.0)
            (0.65,0.5550010894031414) +- (0.0,0.0)
            (0.7000000000000001,0.5569278019415282) +- (0.0,0.0)
            (0.75,0.5589987063398618) +- (0.0,0.0)
            (0.8,0.5601825873607151) +- (0.0,0.0)
            (0.8500000000000001,0.560622858915728) +- (0.0,0.0)
            (0.9,0.5611620561713804) +- (0.0,0.0)
            (0.9500000000000001,0.5643155657216652) +- (0.0,0.0)
            (1.0,0.5649778867714437) +- (0.0,0.0)
    	};
    	\legend{Random, Entropy, Likelihood, Divergence, Agreement, Cluster-Divergence, Cluster-Agreement, Cluster-Random}
	\end{axis}
    \end{tikzpicture}
    \end{subfigure}
    \caption{\small{Validation performance across many potential active learning methods on the MSR-VTT dataset using the transformer model structure with respect to CIDEr Score \cite{vedantam2015cider}, METEOR Score \cite{agarwal2008meteor}, BLEU Score \cite{papineni2002bleu} and ROUGE-L Score \cite{lin2002manual}. The curves presented are the means of 3 individual experiments using each method. Error bars are omitted for clarity. ALISE and Coreset are omitted due to computation time costs (However see Figure \ref{fig:results_main} for a comparison on CIDEr).}}
    \label{fig:results_all}
    \vspace{-1em}
\end{figure}

%% file: tex_figures/cluster_combined.tex
\begin{figure}[t]
    \begin{minipage}{0.48\textwidth}
    \begin{tikzpicture}
    	\begin{axis}[
    		xlabel=Learning Iteration,
    		ylabel=Average distance ($L2$),
    		legend cell align=left,
            legend pos=north east,
            legend style={draw=none, nodes={scale=0.7, transform shape}},
            width=\textwidth,
            height=17em,
    		]
    	\addplot[color=green,mark=x] coordinates {
    	    (0,4.634678021283217)
            (1,4.392140327085192)
            (2,4.292972347865882)
            (3,4.22800517753816)
            (4,4.146207831515153)
            (5,4.113034162003269)
            (6,4.0762952510979575)
            (7,4.052329073009837)
            (8,4.024777004656418)
            (9,3.999790870447754)
            (10,3.978662673136598)
            (11,3.9558982834729632)
            (12,3.9423132964542935)
            (13,3.930844665772958)
            (14,3.920901207856729)
            (15,3.910217500068772)
            (16,3.89363095745954)
            (17,3.890571065591854)
            (18,3.8820668207087987)
            (19,3.878735732264682)
    	};
    	\addplot[color=blue,mark=x] coordinates {
            (0,4.593956397093038)
            (1,4.386666383781663)
            (2,4.272702046803066)
            (3,4.208110740247147)
            (4,4.163797784859026)
            (5,4.1207829294070395)
            (6,4.0950290144569195)
            (7,4.068327405323205)
            (8,4.021503573212345)
            (9,3.9982178623767446)
            (10,3.981537281627386)
            (11,3.964049067295773)
            (12,3.9480248346654943)
            (13,3.939187496960523)
            (14,3.9249691339565715)
            (15,3.9135677804889335)
            (16,3.9067266428734455)
            (17,3.8988719245557575)
            (18,3.88609495105398)
            (19,3.879011949543022)
    	};
    	\addplot[color=red,mark=x] coordinates {
    		(0,4.627238489972274)
            (1,4.501251674993657)
            (2,4.4041751774264055)
            (3,4.351188936463784)
            (4,4.298685085845426)
            (5,4.246795701308989)
            (6,4.2055094654651235)
            (7,4.166819902013245)
            (8,4.144517497997168)
            (9,4.1215728523745625)
            (10,4.101449671885377)
            (11,4.076286479021222)
            (12,4.0545346679342105)
            (13,4.029004925453447)
            (14,4.007367088521271)
            (15,3.983147995332837)
            (16,3.9430973405089658)
            (17,3.912819060281489)
            (18,3.8983982727081483)
            (19,3.878735732264682)
    	};
    	\addplot[color=orange,mark=x] coordinates {
    		(0,4.618422263584866)
            (1,4.374362921570868)
            (2,4.273647563558229)
            (3,4.215016504648466)
            (4,4.170372632427714)
            (5,4.134080622037891)
            (6,4.098744702291201)
            (7,4.062038357828705)
            (8,4.038710394615618)
            (9,4.024932822471174)
            (10,3.9953465173901686)
            (11,3.9784090562125805)
            (12,3.9625582738181713)
            (13,3.944808467772885)
            (14,3.930846395626874)
            (15,3.9166397597468356)
            (16,3.9070870905095183)
            (17,3.894566875588246)
            (18,3.8866929648147983)
            (19,3.878735732264682)
    	};
    	\legend{Clustered-Divergence, Clustered-Agreement, Agreement Only, Random}
    	\end{axis}
    \end{tikzpicture}
    \end{minipage}
    \begin{minipage}{0.48\textwidth}
    \begin{tikzpicture}
    	\begin{axis}[
    		xlabel=\% of Data,
    		ylabel=CIDEr Score,
    		legend cell align=left,
            legend pos=south east,
            legend style={draw=none, nodes={scale=0.7, transform shape}},
            width=\textwidth,
            height=17em,
    		]
    	\addplot[color=green,mark=x] coordinates {
        	(0.05,0.24596979878174607) +- (0.0,0.0)
            (0.1,0.2847399593909266) +- (0.0,0.0)
            (0.15000000000000002,0.31814939732161507) +- (0.0,0.0)
            (0.2,0.32028929578857707) +- (0.0,0.0)
            (0.25,0.33035519803293883) +- (0.0,0.0)
            (0.30000000000000004,0.3323090183871844) +- (0.0,0.0)
            (0.35000000000000003,0.3348876543234065) +- (0.0,0.0)
            (0.4,0.33553859611439446) +- (0.0,0.0)
            (0.45,0.3373427494650068) +- (0.0,0.0)
            (0.5,0.33752770458921877) +- (0.0,0.0)
            (0.55,0.3427363026885376) +- (0.0,0.0)
            (0.6000000000000001,0.3444283998277247) +- (0.0,0.0)
            (0.65,0.344538688634949) +- (0.0,0.0)
            (0.7000000000000001,0.355365885253089) +- (0.0,0.0)
            (0.75,0.3588897094098235) +- (0.0,0.0)
            (0.8,0.35945313302189585) +- (0.0,0.0)
            (0.8500000000000001,0.3598379864020525) +- (0.0,0.0)
            (0.9,0.36823446089666945) +- (0.0,0.0)
            (0.9500000000000001,0.36918954873676235) +- (0.0,0.0)
            (1.0,0.3693006992134557) +- (0.0,0.0)
    	};
    	\addplot[color=blue,mark=x] coordinates {
            (0.05,0.23410126664193612) +- (0.0,0.0)
            (0.1,0.3048497369356562) +- (0.0,0.0)
            (0.15000000000000002,0.32915335274077195) +- (0.0,0.0)
            (0.2,0.3377061843741477) +- (0.0,0.0)
            (0.25,0.343010072846639) +- (0.0,0.0)
            (0.30000000000000004,0.3499812222499056) +- (0.0,0.0)
            (0.35000000000000003,0.35730064529270267) +- (0.0,0.0)
            (0.4,0.3622567107050786) +- (0.0,0.0)
            (0.45,0.3617348663191654) +- (0.0,0.0)
            (0.5,0.3671089315768593) +- (0.0,0.0)
            (0.55,0.368190088244424) +- (0.0,0.0)
            (0.6000000000000001,0.3677068298011455) +- (0.0,0.0)
            (0.65,0.3694207705960271) +- (0.0,0.0)
            (0.7000000000000001,0.36736032545261776) +- (0.0,0.0)
            (0.75,0.3652674814483858) +- (0.0,0.0)
            (0.8,0.3677399376047741) +- (0.0,0.0)
            (0.8500000000000001,0.36855815058563324) +- (0.0,0.0)
            (0.9,0.3720680839191941) +- (0.0,0.0)
            (0.9500000000000001,0.3682937766050682) +- (0.0,0.0)
            (1.0,0.37640532356934814) +- (0.0,0.0)
    	};
    	\addplot[color=black,mark=x] coordinates {
            (0.05,0.24473286956854254) +- (0.0,0.0)
            (0.1,0.28527947680763877) +- (0.0,0.0)
            (0.15000000000000002,0.323063119163559) +- (0.0,0.0)
            (0.2,0.3313242124693308) +- (0.0,0.0)
            (0.25,0.3383184700702595) +- (0.0,0.0)
            (0.30000000000000004,0.3418523990913842) +- (0.0,0.0)
            (0.35000000000000003,0.34668770976561907) +- (0.0,0.0)
            (0.4,0.3451158628665596) +- (0.0,0.0)
            (0.45,0.3461371856925523) +- (0.0,0.0)
            (0.5,0.34821654968836663) +- (0.0,0.0)
            (0.55,0.3488691302674148) +- (0.0,0.0)
            (0.6000000000000001,0.3517585462873762) +- (0.0,0.0)
            (0.65,0.3591649244278588) +- (0.0,0.0)
            (0.7000000000000001,0.3552439664979696) +- (0.0,0.0)
            (0.75,0.36035082597731705) +- (0.0,0.0)
            (0.8,0.37024875258717926) +- (0.0,0.0)
            (0.8500000000000001,0.37116633017451273) +- (0.0,0.0)
            (0.9,0.37133990491307655) +- (0.0,0.0)
            (0.9500000000000001,0.37576697182604745) +- (0.0,0.0)
            (1.0,0.37771785451699146) +- (0.0,0.0)
    	};
    	\addplot[color=red,mark=x] coordinates {
            (0.05,0.24022648211556563) +- (0.0,0.0)
            (0.1,0.28848649109511315) +- (0.0,0.0)
            (0.15000000000000002,0.3117751460666647) +- (0.0,0.0)
            (0.2,0.31397841943997373) +- (0.0,0.0)
            (0.25,0.32655937879453484) +- (0.0,0.0)
            (0.30000000000000004,0.3400974293276734) +- (0.0,0.0)
            (0.35000000000000003,0.34011300137456546) +- (0.0,0.0)
            (0.4,0.3516336363697281) +- (0.0,0.0)
            (0.45,0.3498147698818258) +- (0.0,0.0)
            (0.5,0.35256609548661333) +- (0.0,0.0)
            (0.55,0.35300508147462656) +- (0.0,0.0)
            (0.6000000000000001,0.35035975054490487) +- (0.0,0.0)
            (0.65,0.3516658510014342) +- (0.0,0.0)
            (0.7000000000000001,0.3606722787157409) +- (0.0,0.0)
            (0.75,0.3617212164424207) +- (0.0,0.0)
            (0.8,0.3607774845282108) +- (0.0,0.0)
            (0.8500000000000001,0.36609613176151123) +- (0.0,0.0)
            (0.9,0.36289042801438987) +- (0.0,0.0)
            (0.9500000000000001,0.3689737695805336) +- (0.0,0.0)
            (1.0,0.3734550338082243) +- (0.0,0.0)
    	};
    	\addplot[color=orange,mark=x] coordinates {
            (0.05,0.22687209812697712) +- (0.0,0.0)
            (0.1,0.2698591740476373) +- (0.0,0.0)
            (0.15000000000000002,0.2971479991351384) +- (0.0,0.0)
            (0.2,0.30310000397832015) +- (0.0,0.0)
            (0.25,0.3162711079334679) +- (0.0,0.0)
            (0.30000000000000004,0.3185542339819385) +- (0.0,0.0)
            (0.35000000000000003,0.3274696171542213) +- (0.0,0.0)
            (0.4,0.328080714647497) +- (0.0,0.0)
            (0.45,0.33425927857736887) +- (0.0,0.0)
            (0.5,0.33810355235728073) +- (0.0,0.0)
            (0.55,0.34481945081587706) +- (0.0,0.0)
            (0.6000000000000001,0.34483099779055393) +- (0.0,0.0)
            (0.65,0.34931781898102393) +- (0.0,0.0)
            (0.7000000000000001,0.34984912054196504) +- (0.0,0.0)
            (0.75,0.3505073885871067) +- (0.0,0.0)
            (0.8,0.3591446725357531) +- (0.0,0.0)
            (0.8500000000000001,0.36066753760237946) +- (0.0,0.0)
            (0.9,0.36723520752960537) +- (0.0,0.0)
            (0.9500000000000001,0.3701688847463054) +- (0.0,0.0)
            (1.0,0.37098412179545803) +- (0.0,0.0)
    	};
    	\addplot[color=black,dashed,line width=2pt] coordinates {
            (0.05,0.24830988035500917) +- (0.0,0.0)
            (0.1,0.2783726353055098) +- (0.0,0.0)
            (0.15000000000000002,0.30198405510349546) +- (0.0,0.0)
            (0.2,0.3159627448723251) +- (0.0,0.0)
            (0.25,0.327853738925901) +- (0.0,0.0)
            (0.30000000000000004,0.32597626013776737) +- (0.0,0.0)
            (0.35000000000000003,0.3297829746891937) +- (0.0,0.0)
            (0.4,0.34378034946793923) +- (0.0,0.0)
            (0.45,0.35380126467363304) +- (0.0,0.0)
            (0.5,0.35366740340626224) +- (0.0,0.0)
            (0.55,0.3544366277502148) +- (0.0,0.0)
            (0.6000000000000001,0.3540288727337743) +- (0.0,0.0)
            (0.65,0.36198863882850935) +- (0.0,0.0)
            (0.7000000000000001,0.3590379480407456) +- (0.0,0.0)
            (0.75,0.360732815590431) +- (0.0,0.0)
            (0.8,0.3632944090228063) +- (0.0,0.0)
            (0.8500000000000001,0.368058512002051) +- (0.0,0.0)
            (0.9,0.363444093221117) +- (0.0,0.0)
            (0.9500000000000001,0.3681633971449761) +- (0.0,0.0)
            (1.0,0.3678355115745463) +- (0.0,0.0)
    	};
    	\legend{600 Clusters, 325 Clusters, 200 Clusters, 35 Clusters, 0 Clusters, Random}
    	\end{axis}
    \end{tikzpicture}
    \end{minipage}
    \caption{\small{(Left) Average distance of validation samples to the nearest training sample over the active learning process. Models with improved diversity improve the distance to the training set more rapidly. We suspect this diversity is why random methods work well vs. non-diversity enforced methods as random methods contain a built-in coverage of the dataset. (Right) Performance of the cluster-divergence active learning method across different numbers of clusters. Performance is greater with greater numbers of clusters, until saturation, where performance regresses to random.}}
    \label{fig:cluster_performance}
    \vspace{-1.5em}
\end{figure}

%% file: tex_figures/ensemble_members.tex
\begin{figure}
    \centering
    \begin{tikzpicture}
    	\begin{axis}[
    		xlabel=\% of Data Used,
    		ylabel=CIDEr Score,
    		legend cell align=left,
            legend pos=south east,
            legend style={draw=none},
            width=\textwidth,
            height=13em,
    		]
    	\addplot[color=green,mark=x,error bars/.cd, y dir=both, y explicit] coordinates {
            (0.05, 0.24108208946692108) +- (0.0,0.0)
            (0.1, 0.3193238160907245) +- (0.0,0.0)
            (0.15, 0.34401863008995714) +- (0.0,0.0)
            (0.2, 0.35700901391322064) +- (0.0,0.0)
            (0.25, 0.3630532552327893) +- (0.0,0.0)
            (0.3, 0.37112683726575135) +- (0.0,0.0)
            (0.35, 0.37123265109577336) +- (0.0,0.0)
            (0.4, 0.3732440281873332) +- (0.0,0.0)
            (0.45, 0.3697741583677216) +- (0.0,0.0)
            (0.5, 0.372144032899998) +- (0.0,0.0)
            (0.55, 0.3731849359735934) +- (0.0,0.0)
            (0.6, 0.3747552360327185) +- (0.0,0.0)
            (0.65, 0.37426498937549374) +- (0.0,0.0)
            (0.7, 0.37582833685843885) +- (0.0,0.0)
            (0.75, 0.3753475549254053) +- (0.0,0.0)
            (0.8, 0.37535106331712097) +- (0.0,0.0)
            (0.85, 0.37701134884941267) +- (0.0,0.0)
            (0.9, 0.379179279928347) +- (0.0,0.0)
            (0.95, 0.3832189259408004) +- (0.0,0.0)
            (1.0, 0.3810069223665015) +- (0.0,0.0)
    	};
    	\addplot[color=orange,mark=x,error bars/.cd, y dir=both, y explicit] coordinates {
    		(0.05,0.2326) +- (0.0,0.0)
            (0.1,0.31858) +- (0.0,0.0)
            (0.15,0.33925) +- (0.0,0.0)
            (0.2,0.3503158195165537) +- (0.0,0.0)
            (0.25,0.3619713804040586) +- (0.0,0.0)
            (0.3,0.36974980194252074) +- (0.0,0.0)
            (0.35,0.37022196029791484) +- (0.0,0.0)
            (0.4,0.3694832162562539) +- (0.0,0.0)
            (0.45,0.3696413943267579) +- (0.0,0.0)
            (0.5,0.3698282235223256) +- (0.0,0.0)
            (0.55,0.36402048576683987) +- (0.0,0.0)
            (0.6,0.3634291418218846) +- (0.0,0.0)
            (0.65,0.3690993827602163) +- (0.0,0.0)
            (0.7000000000000001,0.3697310398917909) +- (0.0,0.0)
            (0.75,0.3692674814483858) +- (0.0,0.0)
            (0.8,0.368313531640749) +- (0.0,0.0)
            (0.8500000000000001,0.36855815058563324) +- (0.0,0.0)
            (0.9,0.3720680839191941) +- (0.0,0.0)
            (0.9500000000000001,0.37235969921175987) +- (0.0,0.0)
            (1.0,0.376534954609938) +- (0.0,0.0)
    	};
    	\addplot[color=blue,mark=x,error bars/.cd, y dir=both, y explicit] coordinates {
            (0.05, 0.23171802961054412) +- (0.0,0.0)
            (0.1, 0.3017376108521968) +- (0.0,0.0)
            (0.15, 0.3196052304336207) +- (0.0,0.0)
            (0.2, 0.330363175585666) +- (0.0,0.0)
            (0.25, 0.34247587946272625) +- (0.0,0.0)
            (0.3, 0.34508271025820854) +- (0.0,0.0)
            (0.35, 0.3538668240163165) +- (0.0,0.0)
            (0.4, 0.3572288709886942) +- (0.0,0.0)
            (0.45, 0.36155982047226505) +- (0.0,0.0)
            (0.5, 0.36528443844152875) +- (0.0,0.0)
            (0.55, 0.36682229339886474) +- (0.0,0.0)
            (0.6, 0.36774516583359707) +- (0.0,0.0)
            (0.65, 0.3674248650526493) +- (0.0,0.0)
            (0.7, 0.36827945216461234) +- (0.0,0.0)
            (0.75, 0.3688511098726466) +- (0.0,0.0)
            (0.8, 0.3703522933544757) +- (0.0,0.0)
            (0.85, 0.36830359330901546) +- (0.0,0.0)
            (0.9, 0.37487530760083226) +- (0.0,0.0)
            (0.95, 0.3792787622366768) +- (0.0,0.0)
            (1.0, 0.3752171180233279) +- (0.0,0.0)
    	};
    	\addplot[color=red,mark=x,error bars/.cd, y dir=both, y explicit] coordinates {
    		(0.05,0.24115723123809749) +- (0.013089655530755497,0.0138923862569832)
            (0.1,0.2807601868933432) +- (0.006770766136519769,0.006806213276513492)
            (0.15000000000000002,0.30564417784735065) +- (0.009274445881548568,0.009905951294126398)
            (0.2,0.3127827441303963) +- (0.011843520639098609,0.012391686914742761)
            (0.25,0.325228337520881) +- (0.00724654491933452,0.007527101663859289)
            (0.30000000000000004,0.33502442225016893) +- (0.011268830626727747,0.01149540274116131)
            (0.35000000000000003,0.3295700280101429) +- (0.011912141106795459,0.012109035631855678)
            (0.4,0.3369482079796254) +- (0.005259140315692931,0.00557060999222741)
            (0.45,0.3418220671500454) +- (0.013283717278769529,0.013771626866111561)
            (0.5,0.3448165299451313) +- (0.01071709291236975,0.011054167665195905)
            (0.55,0.35025251853541073) +- (0.004907559340565209,0.004980543202449182)
            (0.6000000000000001,0.35341294598274925) +- (0.0024942954911775605,0.002538206795724208)
            (0.65,0.35749911841918114) +- (0.0077876107031587,0.007904299295982561)
            (0.7000000000000001,0.3551895117223253) +- (0.0041259647668138045,0.004192156849029958)
            (0.75,0.35896006144040504) +- (0.0033199583051125724,0.003611053264438136)
            (0.8,0.36392037899856877) +- (0.006705859608862794,0.007342536882400806)
            (0.8500000000000001,0.37298576626320705) +- (0.0028596857217169785,0.002942924032531293)
            (0.9,0.37346207814992054) +- (0.006940521053014359,0.006984639490820473)
            (0.9500000000000001,0.3738013471945252) +- (0.007688572676029563,0.00790992851056391)
            (1.0,0.3756358775437451) +- (0.0049883162491303375,0.0050648910730082645)
    	};
    	\addplot[color=black, dashed] coordinates {
    	    (0,.357)
    	    (1,.357)
    	} node[anchor=north east] at (axis description cs: 0.22,0.94) {\scriptsize 95\% of Max};
    	\legend{Eight, Four, Two, Random}
    	\end{axis}
    \end{tikzpicture}
    \vspace{-0.5em}
    \caption{\small{Validation performance with differing numbers of ensemble members on the MSR-VTT dataset. We see increasing the number of ensemble members leads to increased performance. We speculate that the diminishing returns are caused by independent models capturing similar information.}}
    \label{fig:ensemble_members}
    \vspace{-2em}
\end{figure}

%% file: tex_figures/lstm_results.tex
\begin{figure}
    \centering
    \begin{tikzpicture}
    	\begin{axis}[
    		xlabel=\% of Data Used,
    		ylabel=CIDEr Score,
    		legend cell align=left,
            legend pos=south east,
            legend style={draw=none, nodes={scale=0.8, transform shape}},
            width=\textwidth,
            height=13em,
    		]
    	\addplot[color=red,mark=x,error bars/.cd, y dir=both, y explicit] coordinates {
            (0.05,0.18323957388816814) +- (0.0,0.0)
            (0.1,0.2014959049376858) +- (0.0,0.0)
            (0.15000000000000002,0.2301948075925525) +- (0.0,0.0)
            (0.2,0.2614881614255601) +- (0.0,0.0)
            (0.25,0.27890550240270706) +- (0.0,0.0)
            (0.30000000000000004,0.2969572308224646) +- (0.0,0.0)
            (0.35000000000000003,0.3137694140305769) +- (0.0,0.0)
            (0.4,0.32786179616087906) +- (0.0,0.0)
            (0.45,0.3310869415484534) +- (0.0,0.0)
            (0.5,0.33552762322406653) +- (0.0,0.0)
            (0.55,0.33399540924465964) +- (0.0,0.0)
            (0.6000000000000001,0.33362372463472655) +- (0.0,0.0)
            (0.65,0.33192944600905616) +- (0.0,0.0)
            (0.7000000000000001,0.3357713476537399) +- (0.0,0.0)
            (0.75,0.32569310679119345) +- (0.0,0.0)
            (0.8,0.3330268825803916) +- (0.0,0.0)
            (0.8500000000000001,0.33822312131495147) +- (0.0,0.0)
            (0.9,0.3381432042765801) +- (0.0,0.0)
            (0.9500000000000001,0.33886091799907714) +- (0.0,0.0)
            (1.0,0.33259866696974344) +- (0.0,0.0)
    	};
    	\addplot[color=black,mark=x,error bars/.cd, y dir=both, y explicit] coordinates {
            (0.05,0.17976614639156702) +- (0.0,0.0)
            (0.1,0.1911639506550827) +- (0.0,0.0)
            (0.15000000000000002,0.198457080108263995) +- (0.0,0.0)
            (0.2,0.220215132400492) +- (0.0,0.0)
            (0.25,0.23619442712887164) +- (0.0,0.0)
            (0.30000000000000004,0.2510627886042612) +- (0.0,0.0)
            (0.35000000000000003,0.2585885623361414) +- (0.0,0.0)
            (0.4,0.2668622259745911) +- (0.0,0.0)
            (0.45,0.3005544256782875) +- (0.0,0.0)
            (0.5,0.31173710493153706) +- (0.0,0.0)
            (0.55,0.31288115004839684) +- (0.0,0.0)
            (0.6000000000000001,0.3292582844769889) +- (0.0,0.0)
            (0.65,0.33006160978826815) +- (0.0,0.0)
            (0.7000000000000001,0.3309775062126714) +- (0.0,0.0)
            (0.75,0.3358989879306903) +- (0.0,0.0)
            (0.8,0.3291412709641142) +- (0.0,0.0)
            (0.8500000000000001,0.336314971335157) +- (0.0,0.0)
            (0.9,0.33320100892093) +- (0.0,0.0)
            (0.9500000000000001,0.3211907149437139) +- (0.0,0.0)
            (1.0,0.3418871096932336) +- (0.0,0.0)
    	};
    	\addplot[color=blue,mark=x,error bars/.cd, y dir=both, y explicit] coordinates {
            (0.05,0.17193287045454575) +- (0.0,0.0)
            (0.1,0.19777363771603206) +- (0.0,0.0)
            (0.15000000000000002,0.22542293389471) +- (0.0,0.0)
            (0.2,0.23973293459012186) +- (0.0,0.0)
            (0.25,0.25271070015612296) +- (0.0,0.0)
            (0.30000000000000004,0.26565140958261687) +- (0.0,0.0)
            (0.35000000000000003,0.2857087708608088) +- (0.0,0.0)
            (0.4,0.29938433393640027) +- (0.0,0.0)
            (0.45,0.304508376025288) +- (0.0,0.0)
            (0.5,0.32247043819218273) +- (0.0,0.0)
            (0.55,0.3336653481867995) +- (0.0,0.0)
            (0.6000000000000001,0.3348500225711918) +- (0.0,0.0)
            (0.65,0.3405632718180606) +- (0.0,0.0)
            (0.7000000000000001,0.345896735221762) +- (0.0,0.0)
            (0.75,0.3454340827496208) +- (0.0,0.0)
            (0.8,0.3427315623071234) +- (0.0,0.0)
            (0.8500000000000001,0.3456161006251856) +- (0.0,0.0)
            (0.9,0.3422989721561126) +- (0.0,0.0)
            (0.9500000000000001,0.34592013751590933) +- (0.0,0.0)
            (1.0,0.34211339836326643) +- (0.0,0.0)
    	};
    	\addplot[color=green,mark=x,error bars/.cd, y dir=both, y explicit] coordinates {
            (0.05,0.16169733309272062) +- (0.0,0.0)
            (0.1,0.17219232893294635) +- (0.0,0.0)
            (0.15000000000000002,0.1892123105188538) +- (0.0,0.0)
            (0.2,0.21228369662921485) +- (0.0,0.0)
            (0.25,0.22651466324892377) +- (0.0,0.0)
            (0.30000000000000004,0.22870241693479796) +- (0.0,0.0)
            (0.35000000000000003,0.25587669176446915) +- (0.0,0.0)
            (0.4,0.25932960041876763) +- (0.0,0.0)
            (0.45,0.2649661066482816) +- (0.0,0.0)
            (0.5,0.27312144107720025) +- (0.0,0.0)
            (0.55,0.2833632256255375) +- (0.0,0.0)
            (0.6000000000000001,0.2955143600528004) +- (0.0,0.0)
            (0.65,0.3001627375508455) +- (0.0,0.0)
            (0.7000000000000001,0.3210548950351478) +- (0.0,0.0)
            (0.75,0.3424507162972493) +- (0.0,0.0)
            (0.8,0.34178112707565195) +- (0.0,0.0)
            (0.8500000000000001,0.34257502733985657) +- (0.0,0.0)
            (0.9,0.34235490222763293) +- (0.0,0.0)
            (0.9500000000000001,0.3425261529668593) +- (0.0,0.0)
            (1.0,0.3420954862089151) +- (0.0,0.0)
    	};
    	\addplot[color=orange,mark=x,error bars/.cd, y dir=both, y explicit] coordinates {
            (0.05,0.1712314055050477) +- (0.0,0.0)
            (0.1,0.1874454390566563) +- (0.0,0.0)
            (0.15000000000000002,0.1986674021821456) +- (0.0,0.0)
            (0.2,0.216947022893178) +- (0.0,0.0)
            (0.25,0.22222931207896175) +- (0.0,0.0)
            (0.30000000000000004,0.2391754566805987) +- (0.0,0.0)
            (0.35000000000000003,0.26528591093656632) +- (0.0,0.0)
            (0.4,0.2692043314396833) +- (0.0,0.0)
            (0.45,0.27403253567154424) +- (0.0,0.0)
            (0.5,0.3122776610553788) +- (0.0,0.0)
            (0.55,0.3230138759990729) +- (0.0,0.0)
            (0.6000000000000001,0.32452677118702085) +- (0.0,0.0)
            (0.65,0.3300934486749413) +- (0.0,0.0)
            (0.7000000000000001,0.33706805686123955) +- (0.0,0.0)
            (0.75,0.3390686047885085) +- (0.0,0.0)
            (0.8,0.3426516683956109) +- (0.0,0.0)
            (0.8500000000000001,0.3439814098057665) +- (0.0,0.0)
            (0.9,0.3414082402273023) +- (0.0,0.0)
            (0.9500000000000001,0.3421939566166063) +- (0.0,0.0)
            (1.0,0.345476981659472) +- (0.0,0.0)
    	};
    	
    	\addplot[color=black, dashed] coordinates {
    	    (0,.3219)
    	    (1,.3219)
    	} node[anchor=north east] at (axis description cs: 0.278,0.94) {\scriptsize 95\% of Max};
    	\legend{Ensemble-Divergence, Random, Coreset, Coreset-Greedy, ALISE}
    	\end{axis}
    \end{tikzpicture}
    \caption{\small{Performance using the LSTM model. While overall performance is lower, the clustered-divergence learning method can save more than 20\% percent of the data.}}
    \label{fig:results_lstm}
\end{figure}

%% file: tex_figures/lsmdc_results.tex
\begin{figure}[t]
    \centering
    \begin{tikzpicture}
    	\begin{axis}[
    		xlabel=\% of Data Used,
    		ylabel=CIDEr Score,
    		legend cell align=left,
            legend pos=south east,
            legend style={draw=none, nodes={scale=0.8, transform shape}},
            width=\textwidth,
            height=13em,
            yticklabel style={
                /pgf/number format/fixed,
                /pgf/number format/precision=5
            },
    		]
    	\addplot[color=red,mark=x,error bars/.cd, y dir=both, y explicit] coordinates {
            (0.05, 0.076)
            (0.1, 0.0817)
            (0.15, 0.0903)
            (0.2,0.09943953693965354) +- (0.0,0.0)
            (0.25,0.1040843473073886) +- (0.0,0.0)
            (0.3,0.11313004229032137) +- (0.0,0.0)
            (0.35,0.11691771175295884) +- (0.0,0.0)
            (0.4,0.1192050761196595) +- (0.0,0.0)
            (0.45,0.1208653840621676) +- (0.0,0.0)
            (0.5000000000000002,0.1225407122392847) +- (0.0,0.0)
            (0.55,0.1292626892185164) +- (0.0,0.0)
            (0.6,0.1266788892953107) +- (0.0,0.0)
            (0.65000000000000004,0.12838107183011) +- (0.0,0.0)
            (0.7000000000000003,0.1281380404727736) +- (0.0,0.0)
            (0.75,0.127819416916582) +- (0.0,0.0)
            (0.8,0.1288325773415245) +- (0.0,0.0)
            (0.85,0.12755869699016535) +- (0.0,0.0)
            (0.9,0.1281501511636375) +- (0.0,0.0)
            (1.0,0.128188527365657) +- (0.0,0.0)
    	};
    	\addplot[color=black,mark=x,error bars/.cd, y dir=both, y explicit] coordinates {
            (0.05,0.07399529769489052) +- (0.0,0.0)
            (0.1,0.07514552632652709) +- (0.0,0.0)
            (0.15000000000000002,0.08181104075675241) +- (0.0,0.0)
            (0.2,0.0917524320586383) +- (0.0,0.0)
            (0.25,0.09232498641170725) +- (0.0,0.0)
            (0.30000000000000004,0.10582744495672211) +- (0.0,0.0)
            (0.35000000000000003,0.10588375279006965) +- (0.0,0.0)
            (0.4,0.11216488629296362) +- (0.0,0.0)
            (0.45,0.11480448287801147) +- (0.0,0.0)
            (0.5,0.11472971051206423) +- (0.0,0.0)
            (0.55,0.11525743364507071) +- (0.0,0.0)
            (0.6000000000000001,0.11711503096612015) +- (0.0,0.0)
            (0.65,0.11715548708512892) +- (0.0,0.0)
            (0.7000000000000001,0.11830029256906903) +- (0.0,0.0)
            (0.75,0.12092741916841645) +- (0.0,0.0)
            (0.8,0.119272669419677501) +- (0.0,0.0)
            (0.8500000000000001,0.12313428886368108) +- (0.0,0.0)
            (0.9,0.12618099114494596) +- (0.0,0.0)
            (0.9500000000000001,0.12861480160218044) +- (0.0,0.0)
            (1.0,0.1276519235321041) +- (0.0,0.0)
    	};
    	\addplot[color=green,mark=x,error bars/.cd, y dir=both, y explicit] coordinates {
            (0.05,0.07160210892805339) +- (0.0,0.0)
            (0.1,0.07253017275124034) +- (0.0,0.0)
            (0.15000000000000002,0.07671241910373902) +- (0.0,0.0)
            (0.2,0.08537853861097633) +- (0.0,0.0)
            (0.25,0.08994720861328256) +- (0.0,0.0)
            (0.30000000000000004,0.09255613124111436) +- (0.0,0.0)
            (0.35000000000000003,0.0998974379649714) +- (0.0,0.0)
            (0.4,0.10395052300947861) +- (0.0,0.0)
            (0.45,0.11101455702879558) +- (0.0,0.0)
            (0.5,0.11522796206428461) +- (0.0,0.0)
            (0.55,0.11503103981727423) +- (0.0,0.0)
            (0.6000000000000001,0.11548019955322199) +- (0.0,0.0)
            (0.65,0.11631003223868232) +- (0.0,0.0)
            (0.7000000000000001,0.1169055617545203) +- (0.0,0.0)
            (0.75,0.11645971727954636) +- (0.0,0.0)
            (0.8,0.11737451152925533) +- (0.0,0.0)
            (0.8500000000000001,0.11801847687419713) +- (0.0,0.0)
            (0.9,0.12494912116182601) +- (0.0,0.0)
            (0.9500000000000001,0.12854244109926993) +- (0.0,0.0)
            (1.0,0.12866770551749863) +- (0.0,0.0)
    	};
    	\addplot[color=orange,mark=x,error bars/.cd, y dir=both, y explicit] coordinates {
            (0.05,0.07156902874417947) +- (0.0,0.0)
            (0.1,0.0807459099246863) +- (0.0,0.0)
            (0.15000000000000002,0.08093999283405722) +- (0.0,0.0)
            (0.2,0.09177093263660664) +- (0.0,0.0)
            (0.25,0.09836874671818186) +- (0.0,0.0)
            (0.30000000000000004,0.10528590671133018) +- (0.0,0.0)
            (0.35000000000000003,0.10997507127855555) +- (0.0,0.0)
            (0.4,0.11396804431551243) +- (0.0,0.0)
            (0.45,0.11586809726534264) +- (0.0,0.0)
            (0.5,0.1178910298851466) +- (0.0,0.0)
            (0.55,0.12072413152851752) +- (0.0,0.0)
            (0.6000000000000001,0.12151493480700616) +- (0.0,0.0)
            (0.65,0.1218624698915831) +- (0.0,0.0)
            (0.7000000000000001,0.12107580412993346) +- (0.0,0.0)
            (0.75,0.12195946319756956) +- (0.0,0.0)
            (0.8,0.1213671486844515) +- (0.0,0.0)
            (0.8500000000000001,0.1266303693948436) +- (0.0,0.0)
            (0.9,0.1284682069869581) +- (0.0,0.0)
            (0.9500000000000001,0.12848678814699716) +- (0.0,0.0)
            (1.0,0.1287888043983583) +- (0.0,0.0)
    	};
    	\addplot[color=black, dashed] coordinates {
    	    (0,.12122)
    	    (1,.12122)
    	} node[anchor=north east] at (axis description cs: 0.28,0.94) {\scriptsize 95\% of Max};
    	\legend{Ensemble-Divergence, Random, Coreset-Greedy, ALISE}
    	\end{axis}
    \end{tikzpicture}
    \caption{\small{Validation performance for the LSMDC dataset. We achieve strong performance using almost 35\% less data. We do not include Coreset, as it took $>24$ hours per active-learning step to compute.}}
    \label{fig:results_lsmdc}
    \vspace{-1.5em}
\end{figure}